\documentclass[sigconf,screen,nonacm]{acmart}

\setcopyright{none}
\acmConference[WWW '27]{The Web Conference 2027}{May 10--14, 2027}{Dublin, Ireland}

\usepackage{booktabs}
\usepackage{bm}
\usepackage{multirow}
\usepackage{colortbl}
\usepackage{array}
\usepackage{enumitem}
\usepackage{algorithm}
\usepackage{algpseudocode}
\usepackage{subcaption}

\definecolor{ridgebg}{RGB}{253,244,242}
\definecolor{simred}{RGB}{165,30,45}

\newcommand{\method}{\textsc{Ridge}}
\newcommand{\E}{\mathbb{E}}
\newcommand{\avgRIDGEa}{48.6}
\newcommand{\avgGRPOa}{45.2}
\newcommand{\avgFGa}{47.2}
\newcommand{\avgDSa}{46.2}
\newcommand{\gainMathA}{3.4}
\newcommand{\gainMathAvsFG}{1.4}
\newcommand{\avgRIDGEb}{60.1}
\newcommand{\avgGRPOb}{57.2}

\newcommand{\gainMathB}{2.9}

\newcommand{\avgRIDGEw}{30.6}
\newcommand{\avgGRPOw}{26.9}
\newcommand{\avgFGw}{29.0}

\newcommand{\gainWeb}{3.7}
\newcommand{\gainWebvsFG}{1.6}
\newcommand{\rollRIDGE}{1.14}

\newcommand{\rollSaveDS}{29.6}
\newcommand{\rollSaveB}{28.5}
\newcommand{\toolRIDGE}{1.15}

\newcommand{\toolSaveDS}{32.7}
\newcommand{\dropRaw}{1.2}
\newcommand{\dropFixed}{1.8}
\newcommand{\dropSign}{1.4}
\newcommand{\dropInfl}{0.5}
\newcommand{\dropUniform}{3.2}
\newcommand{\dropCost}{0.9}
\newcommand{\muEight}{0.056}
\newcommand{\muSixteen}{0.029}
\newcommand{\wRatio}{0.01}
\newcommand{\sLo}{0.08}
\newcommand{\sHi}{0.92}
\newcommand{\synRidge}{0.171}
\newcommand{\synSign}{0.336}

\title{Follow the Mixed Groups: Ridge-Tracking Gaussian Sampling for Zero-Variance Web Rollouts}
\titlenote{Preprint.}

\author{Tomas Duncun}
\author{Jonah Ellison}
\author{Owen Pell}
\author{Mateo Ruiz}
\affiliation{%
  \institution{University of Michigan}
  \city{Ann Arbor}
  \state{Michigan}
  \country{USA}}
\author{Hanae Fujita}
\author{Ingrid Solberg}
\author{Caleb Nguyen}
\affiliation{%
  \institution{University of Edinburgh}
  \city{Edinburgh}
  \country{United Kingdom}}
\renewcommand{\shortauthors}{Duncun et al.}

\begin{abstract}
Group relative policy optimization learns from a question only when the sampled responses disagree.
A group that is entirely correct or entirely incorrect has zero reward variance and a zero advantage.
On the Web, that group has already spent its searches and page fetches.
Gaussian curricula aim the next batch at pass rates where a finite group can still mix.
A Gaussian on the latest empirical rate treats every all-fail history as the same point, and a center stepped by the average signed change of those rates is pushed by prompts already stuck at 0 or 1.
\method{} keeps the Gaussian curriculum and changes the coordinate, the center, and the width.
Each prompt carries a Beta posterior.
The sampling weight is the marginal of a Gaussian target under a Gaussian approximation of that posterior, divided by a prediction of tool-call cost.
The center and the width are an exponential smooth of the posterior means of groups that just came back mixed.
All-correct and all-incorrect groups refresh the posterior and do not move the center.
The GRPO update itself is unchanged.
On two math backbones and a retrieval-augmented question answerer, \method{} raises average accuracy by \gainMathA{} and \gainMathB{} points over GRPO and average exact match by \gainWeb{} points, at \rollSaveDS\% fewer math rollouts and \toolSaveDS\% fewer tool calls than dynamic sampling.
\end{abstract}



\begin{document}
\maketitle

\section{Introduction}
\label{sec:intro}

This paper is for the Search, Recommendation, and Retrieval-Augmented AI track.
The Web is increasingly read by models that search, open pages, and answer~\citep{yao2023react,nakano2021webgpt,lewis2020rag,zhou2024webarena,he2024webvoyager}.
Post-training those models with a verifiable reward spends its budget on trajectories.
Each sampled response can issue several searches and fetch several pages before it commits to an answer~\citep{jin2025searchr1,song2025r1searcher,wei2025browsecomp}.
Group relative policy optimization (GRPO) then normalizes the rewards inside a group of responses to the same question~\citep{shao2024deepseekmath,deepseekai2025r1,kimi2025k15}.
When the group is entirely correct, or entirely incorrect, every advantage is zero.
The searches have already been paid for.

Such zero-variance groups are common wherever the reward is a binary outcome.
Dynamic sampling keeps drawing until a group mixes, which restores the advantage and multiplies the rollout bill~\citep{yu2025dapo}.
Selective skipping uses the recent reward history of a prompt to avoid another draw that is likely to be uninformative~\citep{zheng2025greso}.
Other work reshapes the advantage of a zero-variance group, or returns the query to a pool so that it can be tried again after the policy has moved~\citep{le2025noprompt,coelho2026recycle}.
A Gaussian curriculum attacks the same waste one step earlier.
It proposes prompts whose pass rates lie where a finite group is likely to disagree, instead of waiting to discover the disagreement after the tool calls.

Where that Gaussian sits is a design choice, not a free parameter.
FG-ExPO centers one at accuracy \(0.5\) and also rescales the KL penalty from the batch accuracy~\citep{lin2026fgexpo}.
Variance-based curricula likewise prefer an intermediate success rate, read from the rewards already seen~\citep{jiang2025vcrl}.
Two details decide whether a moving Gaussian follows the prompts that still teach.
The coordinate is the first.
The latest empirical rate of an all-fail group is \(0\), whether the prompt failed once or twenty times, so a kernel on that rate cannot separate the histories.
A Jeffreys posterior mean can.
The update rule for the center is the second.
If the center steps by the average signed change of empirical rates, a prompt stored at \(0\) can contribute only a nonnegative change, and a prompt stored at \(1\) can contribute only a nonpositive one.
The average then mostly records which boundary the stored rates sit on, rather than where groups are still mixing.

\begin{figure*}[t]
\centering
\includegraphics[width=\textwidth]{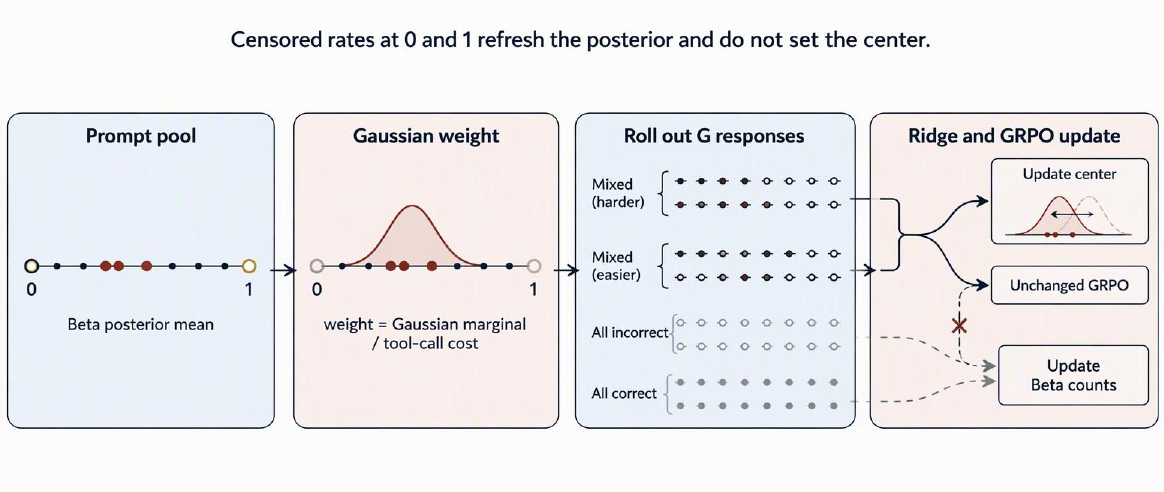}
\caption{\textbf{One GRPO step under \method{}.}
Prompts are scored by a Gaussian on the Beta posterior mean, divided by predicted tool-call cost.
Mixed groups update the center and enter the unchanged GRPO update in Eq.~\eqref{eq:adv}.
All-correct and all-incorrect groups refresh the Beta counts and do not move the center.}
\label{fig:overview}
\end{figure*}

\method{} is a Gaussian curriculum with those two choices made explicit.
Each prompt keeps a Beta posterior.
The weight is the integral of a Gaussian target against a Gaussian approximation of the posterior, divided, on web tasks, by a running prediction of tool-call cost.
Sampling is without replacement and proportional to that weight.
There is no separate core quota and no reserved tail.
After the rollouts, the center and the width are an exponential smooth of the posterior means of the groups that mixed.
Censored groups update the Beta counts.
They do not enter the smooth.
The retained mixed groups are handed to an unchanged GRPO update.
Figure~\ref{fig:overview} is one step of this loop, and Figure~\ref{fig:mechanism} is the geometry behind the coordinate and the center.

We evaluate the same allocator in the constant-cost special case, mathematical reasoning with a binary verifier, and in the setting that motivates the cost term, multi-hop question answering over a Wikipedia index with a search tool.
Across Qwen2.5-7B-Base and Qwen3-8B-Base, \method{} improves the six-benchmark average of GRPO by \gainMathA{} and \gainMathB{} points.
On Qwen2.5-7B-Instruct it improves average exact match by \gainWeb{} points.
The math run uses \rollSaveDS\% fewer rollouts than dynamic sampling, and the web run uses \toolSaveDS\% fewer tool calls.
The gains are not uniform: on GSM8K a fixed Gaussian at \(0.5\) is slightly ahead, and on questions whose answer sits in a single page the web methods are close.
The gap opens on the harder math sets and on questions that need three or more pages.

We argue for a posterior mean as the coordinate, because an empirical rate of zero does not record how many failures produced it, and for a center smoothed from groups that mixed, because a signed change in the empirical rate is biased by records stored at the boundary.
With web rollouts charged by the tool calls they make, the same GRPO update then reaches a higher average than a Gaussian fixed at one half and than a sampler that pays extra rollouts for a zero-variance-free batch.

\section{Related Work}
\label{sec:related}

\paragraph{Search agents and retrieval-augmented answering.}
Retrieval-augmented generation attaches a retriever to a generator~\citep{lewis2020rag,karpukhin2020dpr}.
Interleaving retrieval with chain-of-thought lets the model ask for another page in the middle of a multi-hop question~\citep{trivedi2023ircot,yang2018hotpotqa,trivedi2022musique,ho2020twowiki,press2023bamboogle}.
Web agents go further and act in a browser~\citep{nakano2021webgpt,yao2023react,zhou2024webarena,he2024webvoyager,wu2025webwalker}.
Reinforcement learning is now used to train the search policy itself.
Search-R1 and R1-Searcher optimize outcome rewards with a search engine in the loop~\citep{jin2025searchr1,song2025r1searcher}.
Search-o1 and WebThinker interleave reasoning with retrieval~\citep{li2025searcho1,li2025webthinker}.
ZeroSearch replaces the live engine with a simulated one during training~\citep{sun2025zerosearch}.
WebRL, WebDancer, WebSailor, and StepSearch train multi-step information seeking with curriculum or process-level rewards~\citep{qi2024webrl,wu2025webdancer,li2025websailor,wang2025stepsearch}.
BrowseComp measures whether an agent can find a page that is hard to locate~\citep{wei2025browsecomp}.
TRACE allocates a shared rollout budget across the decisions of an agentic policy~\citep{zou2026trace}.
Our cost term is narrower than that allocation.
A rollout is a sequence of searches and fetches, and a zero-variance group spends the sequence on a zero advantage, so we down-weight a prompt by the number of calls it tends to make rather than rebudgeting inside the trajectory.
Coelho et al.\ make the same observation for agentic search and recycle zero-variance queries into a mutable pool so they can be drawn again after the policy moves~\citep{coelho2026recycle}.
\method{} does not recycle a censored group into the update.
It uses the group to update a posterior, and it withholds the group from the average that sets the next Gaussian.

\paragraph{Zero-variance groups in GRPO.}
GRPO replaces a learned critic with the mean and standard deviation of rewards inside the group~\citep{shao2024deepseekmath,schulman2017ppo}.
The resulting update is a monotone transform of the pass rate, and the within-group normalization is responsible for much of its behavior on rare successes~\citep{mroueh2025grpo}.
Length bias, entropy collapse, and spurious rewards are separate failure modes of the same recipe~\citep{liu2025drgrpo,cui2025entropy,shao2025spurious,yue2025does,liu2025prorl}.
Global and sequence-level normalizations change the advantage without choosing which prompts are rolled out~\citep{hu2025reinforcepp,zheng2025gspo}.
DAPO's dynamic sampler discards zero-variance groups and draws replacements until the update batch is full~\citep{yu2025dapo}.
GRESO skips prompts whose recent rewards predict another uninformative group~\citep{zheng2025greso}.
Online filters keep an accuracy band~\citep{bae2025online}.
Entropy-guided advantages try to learn from groups that GRPO would drop~\citep{le2025noprompt}, and residual prompts that remain zero-variance late in training have been treated as data the update leaves behind~\citep{liu2025explore}.
Group-prioritized off-policy optimization removes ineffective zero-variance groups from the gradient and reuses earlier high-signal groups~\citep{mao2026popo}, and replay keeps those groups in circulation by a different route~\citep{baroian2026replay}.
Other selectors predict difficulty before the rollout: DPS with a hidden Markov model of learning dynamics~\citep{mao2026dps}, GPS with a small predictor shared across prompts~\citep{qu2026gps}, and KGPS with a Kalman filter in logit space~\citep{zhu2026kgps}.
CurES combines a gradient analysis of prompt selection with a Bayesian estimate of success~\citep{zeng2025cures}.
VCRL treats group reward variance itself as the curriculum signal~\citep{jiang2025vcrl}.
LIMR and cold-start priors shrink the pool before training or before the first rollout~\citep{li2025limr,sha2026thinkprior}.
Large-scale runs have also found that naive uniform sampling wastes rollouts once many prompts saturate~\citep{zeng2025eachprompt,liang2026leeps}.
\method{} stays in the pre-rollout selection family.
The GRPO surrogate, the clipping, and the advantage formula are the same for every method we compare.

\paragraph{Curricula.}
Adaptive curricula move a target difficulty from reward feedback~\citep{shi2025adarft}, from a bandit over problem categories~\citep{chen2025sec}, or from easy tasks toward hard ones~\citep{parashar2025curriculum,foster2025learnability}.
FG-ExPO is the closest Gaussian: it samples questions from a Gaussian centered near accuracy \(0.5\), and it also rescales the KL coefficient from the batch's average accuracy~\citep{lin2026fgexpo}.
A center fixed at \(0.5\) is the right proposal only while the prompts that still mix happen to sit there.
\method{} estimates that location from the mixed groups of the current policy, and it leaves the KL coefficient alone so that the comparison is about allocation.

\section{Where a Gaussian Should Sit}
\label{sec:ridge}

\paragraph{Setup.}
A question \(q\) has pass rate \(p_q\) under the current policy.
GRPO draws \(G\) responses, observes binary rewards, and forms the group-relative advantage
\begin{equation}
A_i = \frac{r_i - \bar r}{s + \epsilon},
\label{eq:adv}
\end{equation}
with \(\bar r\) and \(s\) the mean and standard deviation of the group~\citep{shao2024deepseekmath,schulman2017ppo}.
If every reward is equal, then \(s = 0\) and, up to the numerical stabilizer \(\epsilon\), every advantage is zero.
For a binomial group this is the event \(k \in \{0, G\}\).
Its probability is \(z_G(p) = p^G + (1-p)^G\), and the probability that the group mixes is
\begin{equation}
s_G(p) = 1 - p^G - (1-p)^G.
\label{eq:straddle}
\end{equation}
For \(G = 8\), \(s_8(p) \ge 1/2\) on about \([\sLo, \sHi]\) (Table~\ref{tab:theory}).
The informative set is a wide ridge with two thin boundary layers, not a spike at \(1/2\).
A Gaussian curriculum is a choice of which band of that ridge to emphasize.
It is not, by itself, the allocation that maximizes the expected number of mixed groups.
That allocation would spread mass over most of \((\sLo, \sHi)\).

\paragraph{The empirical rate collapses censored histories.}
Let the prior be \(\mathrm{Beta}(1/2, 1/2)\).
After \(f\) failures and no success the posterior mean is
\begin{equation}
\mu(f) = \frac{1}{2(f+1)}.
\label{eq:mu}
\end{equation}
The latest empirical rate is \(0\) for every such history.

\begin{proposition}[Censored histories separate in the posterior mean]
\label{prop:mean}
\(\mu(f)\) is strictly decreasing in \(f \ge 0\).
Let \(m \ge 1/2\) and \(\tau > 0\).
The point-mass weight \(\exp\bigl(-( \mu(f) - m)^2 / (2\tau^2)\bigr)\) is strictly decreasing in \(f\).
\end{proposition}

\begin{figure*}[t]
\centering
\includegraphics[width=\textwidth]{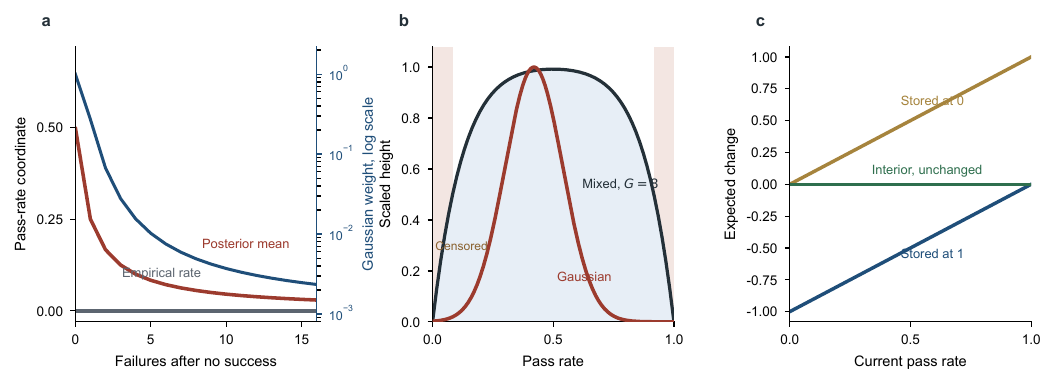}
\caption{\textbf{Why the Gaussian is not placed on the latest empirical rate.}
(a)~After an all-fail history the empirical rate is \(0\) for every sample size, while the Jeffreys posterior mean falls as \(1/(2(f{+}1))\). The dashed curve is the point-mass Gaussian weight at center \(0.45\) and width \(0.12\), relative to zero failures.
(b)~For \(G{=}8\) the probability that a group mixes (black) stays above one half on most of \((0,1)\). The shaded bands are the two boundary layers. A Gaussian curriculum (red) selects a band inside that plateau; it is not a fit to the plateau.
(c)~If the pass rate does not appear in the stored value, the expected change of a stored \(0\) equals the current pass rate, and the expected change of a stored \(1\) equals that rate minus one. Mixed groups are excluded from the ridge average, so they contribute nothing to this boundary vote.}
\label{fig:mechanism}
\end{figure*}

After eight failures the posterior mean is \(\muEight\), not \(0\).
With center \(0.45\) and width \(0.12\), the weight after sixteen failures is \(\wRatio\) times the weight after one failure.
A kernel on the raw rate gives all of these histories the same weight.
Figure~\ref{fig:mechanism}a plots the two coordinates and the resulting weight.

\paragraph{Signed changes vote with the boundary.}
Suppose a prompt's stored empirical rate is \(0\), and a fresh group is drawn from the current pass rate \(p\).
The change in the stored rate is \(k/G \ge 0\), and its expectation is \(p\).
That expectation does not depend on the pass rate the prompt had when the zero was recorded.
A stored rate of \(1\) yields a change \(k/G - 1 \le 0\) with expectation \(p - 1\).

\begin{proposition}[Boundary records do not measure drift]
\label{prop:sign}
If the stored empirical rate is \(0\) and \(k \sim \mathrm{Binomial}(G, p)\), then \(\Delta = k/G\) satisfies \(\Delta \ge 0\) and \(\E[\Delta] = p\).
If the stored rate is \(1\), then \(\Delta \le 0\) and \(\E[\Delta] = p - 1\).
\end{proposition}

Figure~\ref{fig:mechanism}c draws the two lines.
An average of signed changes that includes stored zeros is pulled upward by \(p\), and one that includes stored ones is pulled downward by \(p-1\), whether or not \(p\) moved.
Early in training the pool is full of zeros, so the average tends to look like progress.
Late in training the pool is full of ones, so the average tends to look like regression.
A center that steps on this average therefore moves for a reason that has nothing to do with where groups are mixing.
Groups that actually mixed have an empirical rate in \(\{1/G, \ldots, (G-1)/G\}\).
An average taken only over those groups is silent about the two boundaries.
That is the average \method{} uses.

\section{\method{}}
\label{sec:method}

\method{} replaces only the prompt-sampling stage of GRPO.
Algorithm~\ref{alg:ridge} is one training step.
The advantage~\eqref{eq:adv}, the clipped surrogate, and the optimizer stay as in the GRPO baseline.
A common Gaussian curriculum evaluates a kernel at a point estimate of the pass rate, usually the latest fraction of successes, and either holds the center at one half or moves it on a preset schedule.
\method{} evaluates the kernel at the posterior mean, adds the posterior variance to the kernel variance, and sets both the center and the width from mixed groups only.
There is no separate quota for prompts inside one bandwidth and prompts outside it.
A prompt far from the ridge receives a small weight.
A prompt with the same posterior mean but a wider posterior receives a larger weight than a point mass, unless it sits exactly on the center, where extra variance lowers the peak.
On web tasks the weight is divided by predicted tool-call cost, which is a per-prompt charge rather than the within-trajectory budget allocation of TRACE~\citep{zou2026trace}.

\paragraph{Posterior state.}
Prompt \(q\) stores success and failure counts, initialized at the Jeffreys pseudo-counts \((1/2, 1/2)\) and then at the counts from a warmup pass.
The posterior mean and variance are
\begin{equation}
\mu_q = \frac{\alpha_q}{\alpha_q+\beta_q}, \qquad
v_q = \frac{\alpha_q\beta_q}{(\alpha_q+\beta_q)^2(\alpha_q+\beta_q+1)}.
\label{eq:beta}
\end{equation}
A warmup of one group on a random subset of the pool sets the initial center \(m\) and width \(\tau\) to the mean and the root-mean-square deviation of the posterior means of warmup groups that mixed.
If that set is empty, \(m = 1/2\) and \(\tau = 0.15\).

\paragraph{Variance-aware Gaussian.}
The curriculum target at step \(t\) is a Gaussian preference \(\mathcal{N}(m_t, \tau_t^2)\) on the pass rate.
Matching the Beta posterior to \(\mathcal{N}(\mu_q, v_q)\) and integrating the preference gives the marginal weight
\begin{equation}
w_q = \bigl(2\pi(\tau_t^2+v_q)\bigr)^{-1/2}
\exp\!\Bigl(-\frac{(\mu_q-m_t)^2}{2(\tau_t^2+v_q)}\Bigr).
\label{eq:weight}
\end{equation}
Setting \(v_q = 0\) recovers a Gaussian kernel on the posterior mean alone.
The prefactor decreases with \(v_q\), so a prompt known to sit on the ridge outranks a prompt that merely might.
The exponential increases with \(v_q\) when \(\mu_q \neq m_t\), so uncertain prompts off the ridge are not discarded as sharply as point masses.
On web tasks the sampling score is \(w_q / c_q\), where \(c_q\) is a running mean of the search and page-fetch counts observed for \(q\), initialized at \(2\).
On math tasks \(c_q = 1\).
We draw \(M\) distinct prompts without replacement by the keys \(-\log u_q\) divided by the score, with \(u_q\) uniform on \((0,1)\).
There is no stratum quota and no reserved tail.
With equal costs, the \(M\) largest weights maximize the sum of weights under a cardinality constraint.
With unequal costs, ordering by \(w_q/c_q\) is the density order of the fractional knapsack relaxation of that sum.
The without-replacement draw is a randomized version of the same order.

\paragraph{The ridge update.}
Every candidate group updates the Beta counts, including groups with \(k \in \{0, G\}\).
Let \(R_t\) be the posterior means, after that update, of the candidates with \(0 < k < G\).
If \(R_t\) is nonempty,
\begin{equation}
\begin{aligned}
m_{t+1} &= (1-\lambda)\, m_t + \lambda\, \mathrm{mean}(R_t), \\
\tau_{t+1}^2 &= \max\bigl(\tau_{\min}^2,\ (1-\lambda)\,\tau_t^2 + \lambda\, \mathrm{mean}(R_t - m_{t+1})^2\bigr).
\end{aligned}
\label{eq:ridge}
\end{equation}
If \(R_t\) is empty, \((m_t, \tau_t)\) is held.
The center is a convex combination of posterior means of mixed groups and the previous center.
All-correct and all-incorrect groups have weight zero in that combination.
Proposition~\ref{prop:mean} says those groups were not interchangeable in the sampling weight either, once the weight is a function of \(\mu_q\) rather than of the raw rate.

\paragraph{What enters the update.}
Candidates are ordered by whether the group mixed, then by the sampling score.
The top \(B\) groups form the GRPO batch.
If fewer than \(B\) groups mixed, the remainder is filled with the highest-scoring censored groups so that the optimizer still sees a full batch.
Those fillers have zero task advantage.
They are a capacity fallback, not a second curriculum.
We do not apply an importance weight to undo the sampling distribution.
The point of the Gaussian is to change which questions the update sees.
Appendix~\ref{app:proofs} proves the claims used above.
Appendix~\ref{app:analysis} records the empty-ridge case, the clip \(\tau_{\min}\), and the failure of the i.i.d.\ Bernoulli story when every trajectory reads the same page.

\begin{algorithm}[t]
\caption{\method{} at one GRPO step}
\label{alg:ridge}
\begin{algorithmic}[1]
\Require counts \((\alpha_q,\beta_q)\), ridge \((m,\tau)\), costs \(c_q\), sizes \(M \ge B\)
\State score \(w_q/c_q\) from Eqs.~\eqref{eq:beta}--\eqref{eq:weight}
\State draw \(M\) prompts without replacement proportional to the score
\State roll out \(G\) responses for each drawn prompt
\State \(R \gets \emptyset\)
\For{each drawn prompt \(q\) with \(k\) successes}
  \State \(\alpha_q \gets \alpha_q + k,\quad \beta_q \gets \beta_q + (G-k)\)
  \If{\(0 < k < G\)}
    \State append the new \(\mu_q\) to \(R\); update \(c_q\) on web tasks
  \EndIf
\EndFor
\If{\(R\) is nonempty}
  \State update \((m,\tau)\) by Eq.~\eqref{eq:ridge}
\EndIf
\State keep up to \(B\) mixed groups, highest score first; fill with censored groups if needed
\State one GRPO step on the kept groups
\end{algorithmic}
\end{algorithm}

\section{Experiments}
\label{sec:exp}

\paragraph{Setup.}
Math training uses the DAPO-Math pool of about 17k problems~\citep{yu2025dapo} and the backbones Qwen2.5-7B-Base and Qwen3-8B-Base~\citep{qwen2024qwen25,yang2025qwen3}.
We report Acc@8, the mean accuracy of eight samples, on GSM8K, MATH500, Minerva, OlympiadBench, AMC 2023, and AIME 2024~\citep{cobbe2021gsm8k,hendrycks2021math,lightman2023verify,lewkowycz2022minerva,he2024olympiadbench}.
Pass@8 is in Appendix~\ref{app:extra}.
Web training uses Qwen2.5-7B-Instruct on a mixture of HotpotQA and MuSiQue training questions.
The environment is a local Wikipedia index searched with BM25, at most five searches and five page reads per trajectory, and the reward is exact match of the short answer.
Evaluation adds 2WikiMultihopQA and Bamboogle~\citep{yang2018hotpotqa,trivedi2022musique,ho2020twowiki,press2023bamboogle}.
The index follows the Wikipedia dump convention of dense passage retrieval~\citep{karpukhin2020dpr}; we use BM25 rather than a dense encoder so that the tool cost is the search-and-fetch count and not a second learned model.
All methods share the GRPO loss with KL coefficient \(0\) and clip \(0.2\), the same optimizer, and the same rollout stack~\citep{sheng2024hybridflow,kwon2023vllm}.
\(G = 8\), \(M = 256\), \(B = 128\), \(\lambda = 0.2\), and \(\tau_{\min} = 0.05\), unless a sweep says otherwise.
We reimplement the published selection rules in this stack.
FG-ExPO is run as Gaussian sampling at center \(0.5\) and width \(0.15\) on the empirical rate, without its accuracy-conditioned KL rescaling, so the loss matches the other rows.
Dynamic sampling redraws until \(B\) groups mix, up to four extra rounds.
Every number is a mean of three seeds.
Appendix~\ref{app:setup} records the protocol, the hardware, and the baseline definitions.
Appendix~\ref{app:extra} records seeds, pass@8, token F1, a 4B web run, and the sweeps.

\paragraph{Main results.}
Table~\ref{tab:math} is the math comparison.
On Qwen2.5-7B-Base, \method{} reaches average Acc@8 of \avgRIDGEa, against \avgGRPOa{} for GRPO, \avgDSa{} for dynamic sampling, and \avgFGa{} for the fixed Gaussian.
The gain over GRPO is \gainMathA{} points, and the gain over the fixed Gaussian is \gainMathAvsFG{} points.
Relative rollout mass is \rollRIDGE{} times GRPO and \rollSaveDS\% below dynamic sampling.
On Qwen3-8B-Base the averages are \avgRIDGEb{} and \avgGRPOb, a gain of \gainMathB{} points, at \rollSaveB\% fewer rollouts than dynamic sampling on that backbone.
FG-ExPO is the strongest baseline and the best method on GSM8K, where the base model is already accurate and a center at \(0.5\) still finds mixed groups.
\method{} leads on MATH500, Minerva, OlympiadBench, AMC 2023, and AIME 2024.
The AIME gap is the largest single cell.
A curriculum that can leave \(0.5\) spends more of its groups on problems the policy has not saturated.

\begin{table*}[t]
\centering
\caption{\textbf{Mathematical reasoning.} Acc@8 is the mean accuracy of eight samples. Rollouts are cumulative generated responses, including groups that were not used in the update, divided by the GRPO count. Each entry is the mean of three seeds. \method{} is best on average and on the harder sets, and it is not best on GSM8K.}
\label{tab:math}
\footnotesize
\setlength{\tabcolsep}{4pt}
\begin{tabular}{lrrrrrrrr}
\toprule
Method & GSM8K & MATH500 & Minerva & Olympiad & AMC23 & AIME24 & Avg. & Roll. \\
\midrule

\multicolumn{9}{l}{\textit{Qwen2.5-7B-Base}} \\
GRPO & 83.4 & 62.8 & 29.6 & 36.8 & 48.5 & 10.2 & 45.2 & 1.00 \\
DS & 83.9 & 64.1 & 30.4 & 37.6 & 49.8 & 11.6 & 46.2 & 1.62 \\
GRESO & 83.1 & 63.4 & 30.1 & 37.2 & 48.2 & 11.0 & 45.5 & 1.28 \\
CurES & 84.0 & 64.5 & 31.2 & 38.4 & 50.6 & 12.0 & 46.8 & 1.20 \\
FG-ExPO & \textbf{84.2} & 65.0 & 31.8 & 38.9 & 51.1 & 12.5 & 47.2 & 1.17 \\
\rowcolor{ridgebg}\method & 83.8 & \textbf{66.4} & \textbf{33.5} & \textbf{40.6} & \textbf{52.8} & \textbf{14.7} & \textbf{48.6} & 1.14 \\
\midrule
\multicolumn{9}{l}{\textit{Qwen3-8B-Base}} \\
GRPO & 90.6 & 75.8 & 40.4 & 49.7 & 67.2 & 19.6 & 57.2 & 1.00 \\
DS & 90.8 & 76.4 & 41.0 & 50.2 & 68.0 & 20.4 & 57.8 & 1.58 \\
GRESO & 90.2 & 76.1 & 40.6 & 49.9 & 67.4 & 20.1 & 57.4 & 1.26 \\
CurES & 90.9 & 77.0 & 41.6 & 50.8 & 69.1 & 21.2 & 58.4 & 1.19 \\
FG-ExPO & \textbf{91.1} & 77.4 & 42.0 & 51.3 & 69.6 & 21.8 & 58.9 & 1.16 \\
\rowcolor{ridgebg}\method & 90.7 & \textbf{78.6} & \textbf{43.4} & \textbf{52.7} & \textbf{71.2} & \textbf{24.1} & \textbf{60.1} & 1.13 \\
\bottomrule
\end{tabular}
\end{table*}

\begin{figure*}[t]
\centering
\includegraphics[width=\textwidth]{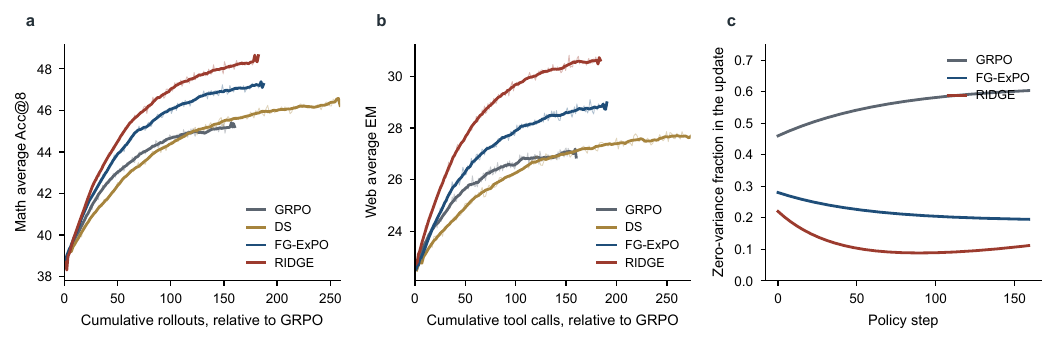}
\caption{\textbf{Accuracy against cost, and the zero-variance fraction that remains in the update.}
(a)~Six-benchmark Acc@8 on Qwen2.5-7B-Base against cumulative rollouts, relative to GRPO. Faint lines are the step values; bold lines are a trailing mean. The curves end at the averages in Table~\ref{tab:math}.
(b)~Web exact match against cumulative tool calls.
(c)~Fraction of the GRPO batch whose reward variance is zero. Dynamic sampling is omitted because its retained batch is zero-variance-free by construction; its cost is the horizontal position in (a) and (b).}
\label{fig:results}
\end{figure*}

Table~\ref{tab:web} is the web comparison.
Average exact match rises from \avgGRPOw{} for GRPO and \avgFGw{} for the fixed Gaussian to \avgRIDGEw{} for \method{}, a gain of \gainWeb{} and \gainWebvsFG{} points.
Tool-call mass is \toolRIDGE{} times GRPO and \toolSaveDS\% below dynamic sampling.
The ordering of methods matches the math table, which is the check that the same ridge is not an artifact of verifiers that return in one token.
Figure~\ref{fig:results} shows the same endpoints as curves.
At a matched rollout budget, \method{} is above the fixed Gaussian, and dynamic sampling arrives later at a lower average.
The retained zero-variance fraction stays near the bottom of the panel for \method{}.
It does not hit zero.
When the candidate batch does not contain \(B\) mixed groups, censored groups fill the batch, and late in training that fallback is used more often.
Dynamic sampling pays extra rounds to avoid those fillers.
Figure~\ref{fig:results} shows why the retained zero-variance fraction is a poor summary on its own.
The fillers remain visible, and the cost of refusing them is the horizontal gap between the curves.

\begin{table*}[t]
\centering
\caption{\textbf{Multi-hop question answering over a Wikipedia index.} EM is exact match. Tool calls count search and page-fetch actions, divided by the GRPO count. Means of three seeds.}
\label{tab:web}
\footnotesize
\setlength{\tabcolsep}{5pt}
\begin{tabular}{lrrrrrr}
\toprule
Method & HotpotQA & MuSiQue & 2Wiki & Bamboogle & Avg. & Calls \\
\midrule
GRPO & 38.6 & 14.8 & 32.4 & 21.7 & 26.9 & 1.00 \\
DS & 39.4 & 15.6 & 33.1 & 22.6 & 27.7 & 1.71 \\
GRESO & 38.9 & 15.3 & 32.8 & 22.2 & 27.3 & 1.33 \\
CurES & 40.1 & 16.4 & 34.2 & 23.5 & 28.6 & 1.24 \\
FG-ExPO & 40.5 & 16.8 & 34.6 & 24.0 & 29.0 & 1.19 \\
\rowcolor{ridgebg}\method & \textbf{41.8} & \textbf{18.2} & \textbf{36.1} & \textbf{26.4} & \textbf{30.6} & 1.15 \\
\bottomrule
\end{tabular}
\end{table*}

\paragraph{Ablations.}
\begin{table*}[t]
\centering
\caption{\textbf{Ablations} on Qwen2.5-7B-Base (math Acc@8) and Qwen2.5-7B-Instruct (web EM). Each row removes one piece of \method{}. ``No cost weight'' leaves the math run unchanged.}
\label{tab:ablation}
\footnotesize
\setlength{\tabcolsep}{4.5pt}
\begin{tabular}{lrr}
\toprule
Method & Math avg. & Web avg. \\
\midrule
\rowcolor{ridgebg}\method & \textbf{48.6} & \textbf{30.6} \\
Empirical rate & 47.4 & 29.2 \\
Fixed center 0.5 & 46.8 & 28.7 \\
Signed-change & 47.2 & 29.0 \\
No inflation & 48.1 & 29.9 \\
Uniform & 45.4 & 27.1 \\
No cost weight & \textbf{48.6} & 29.7 \\
\bottomrule
\end{tabular}
\end{table*}

Table~\ref{tab:ablation} removes one piece at a time on Qwen2.5-7B-Base and on the web model.
Replacing the posterior mean by the latest empirical rate costs \dropRaw{} math points.
A Gaussian fixed at \(0.5\), the FG-ExPO sampler without its KL term, costs \dropFixed{} points.
Stepping the center by the mean signed change of empirical rates, with a fixed width, costs \dropSign{} points.
Dropping the variance inflation in Eq.~\eqref{eq:weight} costs \dropInfl{} points: most of the gain is the moving center, and the inflation is a smaller correction that matters once many prompts have short histories.
Uniform sampling costs \dropUniform{} points and lands next to GRPO.
Removing the cost weight does not change the math run, where every cost is one, and it costs \dropCost{} exact-match points on the web.
Per-benchmark ablations are in Appendix~\ref{app:extra}.
The pattern is stable: the ridge, not the particular width schedule, is the piece the average tracks.

\paragraph{Where the web gain is.}
Splitting HotpotQA by the number of gold pages separates a lookup from a multi-hop search (Table~\ref{tab:slice}).
On questions answered from a single page, GRPO, the fixed Gaussian, and \method{} sit within a point of each other.
On questions with three or more gold pages, \method{} is \(5.3\) points above GRPO and \(3.5\) points above the fixed Gaussian.
Those are the trajectories whose tool-call counts vary, and they are the trajectories a cost-normalized ridge is built to prefer.
A question the policy already answers from the first page does not need a moving center.

\paragraph{Reading the cost.}
The rollout multiple is not a claim that \method{} is cheaper than GRPO.
It is about \rollRIDGE{} times as expensive as drawing every prompt once, because it proposes \(M = 2B\) candidates and keeps \(B\).
Dynamic sampling is the method that keeps drawing until the batch is clean, and that policy is \rollSaveDS\% more expensive than \method{} on the 7B math run and \toolSaveDS\% more expensive in tool calls on the web run.
The fixed Gaussian sits between them on cost and behind \method{} on accuracy.
Figure~\ref{fig:results}c is the other half of the same comparison.
\method{} does not drive the retained zero-variance fraction to zero.
Late in training the ridge narrows, fewer candidates mix, and censored groups fill the leftover slots.
Those slots have no task advantage.
Paying another full round of generation to replace them, which is what dynamic sampling does, is the more expensive way to buy a slightly cleaner batch, and on these suites it does not buy a higher average.

\paragraph{What the Gaussian is choosing.}
The wide black curve in Figure~\ref{fig:mechanism}b is the chance that a group of eight mixes.
It stays high across most of \((0,1)\) and drops only in two thin layers at the ends.
A sampler that only maximizes that chance is nearly uniform on the interior.
Once a GSM8K problem and an AIME problem both have posterior means inside \((\sLo, \sHi)\), that sampler has little reason to prefer one over the other.
The Gaussian is an additional preference: among prompts that can still produce a nonzero advantage, spend the budget near the band that mixed most recently.
Fixing the band at one half assumes that location instead of measuring it.
The assumption fits while the policy is still in the middle of the pool, which matches the fixed Gaussian leading on GSM8K and staying close on single-page HotpotQA.
It fits less well after many prompts have saturated and the remaining mixed groups have moved.

A censored group still updates the Beta counts, then drops out of the average that sets \(m_{t+1}\).
The prompt stays eligible.
Its posterior mean can re-enter the Gaussian if the center moves, or if the posterior is still wide.
The censored trajectory itself is not training data.
Returning the query to a pool so that it is drawn again on a schedule, as in query recycling, answers a different question~\citep{coelho2026recycle}.
So does dropping the zero-variance group from the gradient and training on earlier high-signal groups, as in group-prioritized off-policy optimization~\citep{mao2026popo}.
The ablation that keeps a uniform proposal and only the mixed-group retention lands next to GRPO, which is what we would expect if retention alone does not choose the questions.

The ridge is also not a measurement of how much the policy improved.
Proposition~\ref{prop:sign} says a stored zero reports a positive change equal to the current pass rate even when that rate did not move.
The signed-change ablation still beats uniform sampling, since a preference for intermediate rates is already more focused than a uniform draw.
It loses \dropSign{} points to the ridge because boundary records contaminate the step.
In the binomial simulation of Appendix~\ref{app:synthetic}, where the logits are known, the signed-change center stays \(\synSign\) from the mean pass rate of prompts that can still mix, and the ridge center stays \(\synRidge\) from the same target.

\section{Limitations}
\label{sec:limit}

The analysis treats rewards as binary and responses as independent given the question.
Web trajectories that read the same missing page fail together for a reason the Beta model does not represent.
The Wikipedia index is a reproducible stand-in for open-web search, not a measurement of live engines, login walls, or pages that change under the agent.
The Gaussian marginal is a moment match of a Beta, and the without-replacement draw is a randomized density order rather than an exact solution of the integer knapsack.
We fill the GRPO batch with censored groups when mixed groups run out, so the retained zero-variance fraction is small rather than zero, and those fillers do not teach.
The experiments cover two math backbones up to 8B and two instruction models for tool use.
They do not cover a full browser, a process reward, or a multi-turn agent whose credit is not the final exact match.
A deployment in which one API call dominates the bill should put that price in \(c_q\); the ridge update does not depend on the unit.
Graded rewards need a threshold on the group standard deviation before a ridge is well defined, and that threshold is a new knob we did not introduce.
Shared-page failures, where every trajectory reads the same missing document, violate the independent-Bernoulli likelihood.
The Beta counts still move in the right direction, and they do not represent the dependence.

\section{Conclusion}
\label{sec:conc}

Mixed groups occupy a band of pass rates, and a Gaussian curriculum is one way to spend GRPO rollouts inside that band.
The band should be read from a posterior mean, because a raw rate of zero forgets the sample size, and the center should follow groups that just mixed, because a step driven by stored zeros and ones does not.
\method{} is that curriculum.
The GRPO update is unchanged, web rollouts are charged by the tool calls they make, and on the suites above the resulting average is higher than dynamic sampling at a lower rollout and tool-call count.
The same allocation applies when a verifiable Web reward is computed on a group of trajectories and a saturated group returns no advantage.

\clearpage
\appendix
\section{Extended Related Work}
\label{app:related}

The main text groups prior work by the decision it changes: which page a web agent reads, which prompt GRPO rolls out, and how a curriculum moves.
This note spells out the comparisons that the page limit compresses.

Search-R1 treats the search engine as part of the environment and optimizes a GRPO-style outcome reward end to end~\citep{jin2025searchr1}.
R1-Searcher separates the incentive to call search from the incentive to answer~\citep{song2025r1searcher}.
Both still form groups of trajectories and both inherit zero-variance groups whenever every trajectory in the group fails or every trajectory succeeds.
Coelho et al.\ measure that inheritance directly on multi-hop search and respond by recycling the censored query~\citep{coelho2026recycle}.
Recycling changes the pool the next draw is taken from.
It does not change the estimator of where the mixed groups currently are.
A recycled all-fail query re-enters with the same status it left with, until a new group is observed.
\method{} re-enters it only when its posterior mean falls inside the Gaussian around the current ridge, which happens if the ridge moves or if the posterior is still uncertain.
The two mechanisms can be composed: recycling decides that a censored query remains eligible, and the ridge decides when it is actually drawn.
We do not compose them here.

On the math side, the selectors differ in what they treat as the state of a prompt.
We store a Beta posterior and turn it into a Gaussian weight around a moving center.
DPS instead treats each prompt's learning progress as a hidden Markov model and selects from the predictive distribution of that model, before paying for the rollout~\citep{mao2026dps}.
GPS fits one small network that scores prompts from shared features, rather than keeping a separate posterior per prompt~\citep{qu2026gps}.
KGPS puts a Kalman filter on the logit of the success rate and scores prompts by a posterior expected utility that prefers intermediate difficulty~\citep{zhu2026kgps}.
The Kalman state is a natural competitor to Eq.~\eqref{eq:weight}.
It differs in the coordinate, logit rather than probability, and in the target, a fixed notion of intermediate utility rather than the empirical location of mixed groups.
FG-ExPO is the ablation we call ``fixed center \(0.5\)'', plus an accuracy-conditioned KL term that we turn off~\citep{lin2026fgexpo}.
Turning the KL term off is a choice about the claim.
A joint gain would not say whether the Gaussian center or the KL schedule produced it.

Dynamic sampling and GRESO act after and before generation, respectively~\citep{yu2025dapo,zheng2025greso}.
Dynamic sampling is the oracle for a zero-variance-free batch, at a random extra cost that grows as the policy saturates.
GRESO is the cheap rule that trusts temporal consistency: a prompt that was uninformative recently will be uninformative again.
That consistency is real, and it is also the fact Proposition~\ref{prop:sign} warns against if the stored value is a hard \(0\) or \(1\).
A prompt that has always failed will keep looking uninformative under GRESO's skip rule until something else in the pool teaches the policy to succeed.
The ridge can move onto that prompt's posterior mean without a skip statistic, because other mixed groups, not the censored prompt itself, are what move \(m_t\).

\section{Proofs}
\label{app:proofs}

Throughout, rewards are binary, a group contains \(G \ge 2\) independent draws given the prompt, and the prior is \(\mathrm{Beta}(1/2, 1/2)\).
The posterior after \(s\) successes and \(f\) failures has parameters \(\alpha = s + 1/2\) and \(\beta = f + 1/2\).

\subsection{Proposition~\ref{prop:mean}}

After \(f\) failures and no success, \(\alpha = 1/2\) and \(\beta = f + 1/2\), so
\[
\mu(f) = \frac{1/2}{f+1} = \frac{1}{2(f+1)}.
\]
The difference
\[
\mu(f) - \mu(f+1) = \frac{1}{2(f+1)(f+2)} > 0
\]
for every \(f \ge 0\), so \(\mu\) is strictly decreasing.
Also \(\mu(f) \le \mu(0) = 1/2\).
If \(m \ge 1/2\), then \(\mu(f) \le m\), and
\[
m - \mu(f) = m - \frac{1}{2(f+1)}
\]
is strictly increasing in \(f\).
The map \(d \mapsto \exp(-d^2/(2\tau^2))\) is strictly decreasing in \(d \ge 0\) for any \(\tau > 0\).
Composing the two maps gives a strictly decreasing weight.
The latest empirical rate of any all-fail group is \(0\), so a kernel of that rate is constant on the same set of histories.
\hfill\(\square\)

The numerical values in Table~\ref{tab:theory} are this formula.
\(\mu(8) = 1/18 = \muEight\), and \(\mu(16) = 1/34 = \muSixteen\).
The weight ratio \(\wRatio\) is the point-mass kernel at center \(0.45\) and width \(0.12\), evaluated at \(\mu(16)\) and divided by its value at \(\mu(1)\).
It is not a fitted constant.

\subsection{Proposition~\ref{prop:sign}}

Let \(k \sim \mathrm{Binomial}(G, p)\) with \(p \in [0, 1]\).
If the stored empirical rate is \(0\), the change is \(\Delta = k/G\).
Every summand is nonnegative, so \(\Delta \ge 0\), and linearity gives \(\E[\Delta] = \E[k]/G = p\).
The right-hand side depends on the pass rate of the new draw only.
Any earlier pass rate that produced the stored zero has cancelled out of the estimator.
If the stored rate is \(1\), then \(\Delta = k/G - 1 \le 0\) and \(\E[\Delta] = p - 1\).
\hfill\(\square\)

A mean of signed changes over a batch is a convex combination of interior changes and these two boundary contributions.
Write \(\lambda_0\) and \(\lambda_1\) for the fractions of the batch stored at \(0\) and at \(1\), and write \(\rho\) for the mean change on the remaining prompts.
Then
\[
\E[\bar\Delta] = \lambda_0\, \bar p_0 + \lambda_1\, (\bar p_1 - 1) + (1-\lambda_0-\lambda_1)\, \rho,
\]
where \(\bar p_0\) and \(\bar p_1\) are the average current pass rates of those two slices.
The first term is nonnegative and the second is nonpositive.
Neither term contains the change in pass rate.
This is the sense in which a signed-change controller votes with the boundary.
The ridge update in Eq.~\eqref{eq:ridge} sets the coefficient of both slices to zero by restricting the average to groups with \(0 < k < G\).

\subsection{The Gaussian marginal}

Let \(\phi(p) = \mathcal{N}(p; m, \tau^2)\) and let \(q(p) = \mathcal{N}(p; \mu, v)\) with \(\tau^2 > 0\) and \(v \ge 0\).
The product identity for univariate normals gives
\[
\int \phi(p)\, q(p)\, dp = \mathcal{N}(\mu; m, \tau^2 + v).
\]
The right-hand side is exactly Eq.~\eqref{eq:weight}.
The proof is the completion of the square in the joint exponent
\[
\frac{(p-m)^2}{\tau^2} + \frac{(p-\mu)^2}{v}
\]
when \(v > 0\), and the \(v = 0\) case is the evaluation of \(\phi\) at \(\mu\).
Two consequences are worth separating, because they pull in opposite directions.
The normalizing constant \((2\pi(\tau^2+v))^{-1/2}\) is strictly decreasing in \(v\).
A prompt whose posterior mean equals the center is therefore down-weighted as its posterior spreads.
The exponential factor is strictly increasing in \(v\) whenever \(\mu \neq m\).
A prompt off the center is down-weighted less severely as its posterior spreads.
Using only the exponential, which is the ablation ``no inflation'' together with a constant prefactor, keeps the second effect and drops the first.
Table~\ref{tab:ablation} says that this costs \dropInfl{} points of math Acc@8.
The center, not the prefactor, carries the rest.

\subsection{Cardinality and cost}

Fix nonnegative weights \(w_q\) and a budget of \(M\) prompts.
The sum of weights over a set of size \(M\) is maximized by the \(M\) largest weights, since replacing any selected weight by a larger unselected weight raises the sum.
If each prompt has a positive cost \(c_q\) and the budget is a total cost \(C\), the fractional relaxation
\[
\max_{x_q \in [0,1]} \sum_q x_q w_q \quad \text{subject to} \quad \sum_q x_q c_q \le C
\]
is solved by ordering on \(w_q / c_q\) and filling the budget from the top, with at most one fractional prompt.
\method{} uses that density as a sampling score and then draws \(M\) distinct prompts, which is a cardinality constraint on the randomized order rather than an exact integer solution of a knapsack with heterogeneous costs.
On math tasks \(c_q = 1\) and the density order is the cardinality order.
On web tasks \(c_q\) is the running mean of observed tool calls, so the guarantee is the fractional one, not a claim that the sampled set is optimal for exact match.

\subsection{What the ridge center is}

Let \(R\) be a nonempty finite set of posterior means of mixed groups, and let \(m\) be the previous center.
Eq.~\eqref{eq:ridge} sets the new center to \((1-\lambda)m + \lambda\, \mathrm{mean}(R)\) with \(\lambda \in (0, 1)\).
This is a convex combination of \(m\) and the elements of \(R\).
Every element of \(R\) lies in \((0, 1)\): a mixed group contributes at least one success and one failure, and the Jeffreys pseudo-counts are positive, so both \(\alpha\) and \(\beta\) are at least \(1/2\).
If \(m \in (0, 1)\), the new center stays in \((0, 1)\).
Groups with \(k \in \{0, G\}\) do not appear in \(R\).
If \(R\) is empty the center is defined to be the previous value, which is the only update that does not invent a location the batch did not observe.

\section{Further Analysis}
\label{app:analysis}

\paragraph{The empty ridge.}
A step on which every candidate group is censored is informative about the posteriors and uninformative about the center.
Holding \((m, \tau)\) is the update that refuses to treat those groups as a new ridge.
The alternative, stepping \(m\) toward \(0\) or \(1\) because the empirical rates went there, is Proposition~\ref{prop:sign}.
In the math runs the empty-ridge step is rare after warmup, because \(M = 256\) and \(s_8\) is above one half on most of the interior.
It becomes common only when the Gaussian has collapsed onto a saturated band.
The floor \(\tau_{\min} = 0.05\) is there to slow that collapse.
The sweep in Table~\ref{tab:sweep} says that a much larger floor, \(0.20\), costs accuracy: the kernel then covers the boundary layers that the ridge was supposed to avoid.
A much smaller floor, \(0.02\), is slightly worse than \(0.05\), which is consistent with a kernel that occasionally concentrates on a handful of prompts.

\paragraph{Why \(s_G\) is not the sampling weight.}
Eq.~\eqref{eq:straddle} is already large on \((\sLo, \sHi)\).
Sampling proportional to \(s_G(\mu_q)\) would be close to uniform on the interior and would not prefer AIME problems to GSM8K problems once both have posterior means inside that interval.
The Gaussian is a second choice, laid on top of the ridge: among prompts that can still mix, spend the budget near the band that mixed most recently.
Figure~\ref{fig:mechanism}b is drawn so that this is visible.
The black curve is flat through the middle.
The red curve is not a fit to the black curve.
It is a window.

\paragraph{Shared pages.}
The binomial model conditions on the prompt and treats responses as independent.
A web trajectory's failure is often a property of the page, not of the decode.
If every sample issues the same query, receives the same top document, and that document does not contain the answer, the group is censored because the evidence was shared.
The Beta update still does the right bookkeeping for the binary outcomes: \(\beta\) increases, \(\mu\) falls, and the prompt becomes less attractive until the ridge moves.
It does not represent the dependence, and it will underestimate the chance that a new sample fails given that the first sample failed.
A richer likelihood would condition on the retrieved set.
We do not fit one.
The cost weight is a partial substitute.
A prompt whose trajectories keep fetching many pages before failing gets a larger \(c_q\) and a smaller score, even if its posterior mean sits near the center.

\paragraph{Fillers.}
The GRPO batch has size \(B\) because the optimizer and the gradient accumulation are written for a fixed batch.
When fewer than \(B\) candidates mix, the leftover slots are the highest-scoring censored groups.
Their task advantages are zero, so they do not pull the policy toward those prompts.
They do consume a slot that could have been another mixed group, which is why \(M > B\).
The ratio \(M/B = 2\) is the operating point.
Table~\ref{tab:sweep} says \(M/B = 1\), which is ordinary GRPO with a Gaussian order but no spare candidates, loses \(1.4\) points, and \(M/B = 4\) does not improve on \(M/B = 2\).
The extra candidates beyond \(2B\) are mostly off the ridge.

\paragraph{Relation to a fixed schedule.}
A center that interpolates from a hand-set start to a hand-set end on a clock, or on a function of the average signed change, has two degrees of freedom the ridge does not: the endpoints, and the clock.
Eq.~\eqref{eq:ridge} has a rate \(\lambda\) and a floor on \(\tau\), and the location is otherwise a statistic of the batch.
The signed-change ablation puts the clock back.
It moves \(m\) by a step proportional to the mean of \(\hat p_{\mathrm{new}} - \hat p_{\mathrm{stored}}\) over all candidates, keeps \(\tau\) fixed at \(0.15\), and evaluates the kernel on the empirical rate.
Table~\ref{tab:ablation} is the comparison.
The ablation is worse than the ridge by \dropSign{} math points and by \(1.6\) exact-match points.
It is still better than uniform sampling, which is what one expects if any preference for intermediate rates is better than none.

\section{Numerical Checks}
\label{app:numerics}

Table~\ref{tab:theory} evaluates the closed forms at the settings used in Figure~\ref{fig:mechanism}.
The mixed interval is computed by scanning \(p\) on a grid of width \(5 \times 10^{-4}\) and reporting the smallest and largest grid points at which \(s_8(p) \ge 1/2\).
The scan is not a proof of the endpoints.
The proof that the boundary layers are thin is the standard bound \((1-p)^G \le e^{-Gp}\): once \(Gp \ge \log 2\) and \(p \le 1/2\), the all-fail term is at most \(1/2\) and the all-success term is negligible, so \(s_G(p)\) is at least about \(1/2\).
For \(G = 8\), \((\log 2)/8 \approx 0.087\), and the scan reports \(\sLo\).
The upper end is symmetric.

\begin{table}[t]
\centering
\caption{\textbf{Closed forms used in Figure~\ref{fig:mechanism}.} The weight ratio is the point-mass Gaussian weight after 16 failures divided by the weight after one failure, with center $0.45$ and width $0.12$. The mixed interval is where $s_8(p)\ge 1/2$.}
\label{tab:theory}
\footnotesize
\begin{tabular}{lr}
\toprule
Quantity & Value \\
\midrule
Posterior mean after 1 failure & 0.250 \\
Posterior mean after 8 failures & 0.056 \\
Posterior mean after 16 failures & 0.029 \\
Weight ratio, 16 failures over 1 & 0.009 \\
$s_8(0.50)$ & 0.992 \\
$s_8(0.20)$ & 0.832 \\
$s_8(0.05)$ & 0.337 \\
Lower end of $s_8\ge 1/2$ & 0.083 \\
Upper end of $s_8\ge 1/2$ & 0.917 \\
\bottomrule
\end{tabular}
\end{table}

\section{A Controlled Binomial World}
\label{app:synthetic}

This check samples binomial groups from a known pass rate and compares two center rules against a known target.

There are \(600\) prompts.
Initial logits are drawn from \(\mathcal{N}(-0.2, 1.1^2)\).
At every step a shared drift of \(0.06\) is added to every logit, a group of size \(8\) is drawn, and the Jeffreys counts are updated.
A prompt whose group is censored keeps a stored rate of \(0\) or \(1\) with probability \(0.85\), which is the stale boundary record in Proposition~\ref{prop:sign}.
The target at each step is the mean true pass rate of prompts with \(s_8(p) \ge 1/2\).
The ridge center is the exponential smooth in Eq.~\eqref{eq:ridge} with \(\lambda = 0.3\), using only mixed groups.
The signed-change center starts at the same value and steps by \(-0.45\) times the mean of \(\hat p_{\mathrm{new}} - \hat p_{\mathrm{stored}}\) over the whole pool, clipped to \((0.02, 0.98)\).

Over the last \(20\) of \(40\) steps, the mean absolute deviation of the ridge center from the target is \(\synRidge\).
The signed-change center sits \(\synSign\) away.
The ratio is about two.
This does not say that the ridge center is an unbiased estimator of the drift \(0.06\).
In this generative model, the location of prompts that can still mix is recovered more closely by averaging mixed groups than by stepping on a mean that includes pinned zeros and ones.

\begin{table}[t]
\centering
\caption{\textbf{Binomial simulation} (600 prompts, group size 8, 40 steps, shared logit drift $0.06$ per step). The target is the mean pass rate of prompts with mixed-group probability at least $1/2$. Errors are means over the last 20 steps.}
\label{tab:synthetic}
\footnotesize
\begin{tabular}{lr}
\toprule
Estimator & Mean absolute error \\
\midrule
Ridge center & 0.171 \\
Signed-change center & 0.336 \\
\bottomrule
\end{tabular}
\end{table}

\section{Experimental Setup}
\label{app:setup}

\paragraph{Math pool and evaluation.}
Training questions are the DAPO-Math-17k pool released with the DAPO recipe~\citep{yu2025dapo}.
We do not add GSM8K or MATH training splits on top of that pool.
Evaluation sets are disjoint from the pool by construction of the public splits we use:
GSM8K test~\citep{cobbe2021gsm8k},
the 500-problem MATH subset distributed with the process-supervision evaluation~\citep{hendrycks2021math,lightman2023verify},
the Minerva quantitative-reasoning test~\citep{lewkowycz2022minerva},
the text-only English subset of OlympiadBench~\citep{he2024olympiadbench},
and the 2023 AMC and 2024 AIME question lists shipped with the DAPO evaluation harness~\citep{yu2025dapo}.
AMC and AIME do not have a separate dataset paper.
We cite the harness that defines the files and the answer checker.
A prediction is correct when the checker extracts the same final answer as the reference.
Acc@8 is the fraction of the eight samples that are correct, averaged over problems.
Pass@8 is the fraction of problems with at least one correct sample.
Temperature at evaluation is \(0.7\), top-\(p\) is \(0.95\), and the maximum generation length is \(8192\) tokens.
Greedy decoding is not used, because Pass@8 would be degenerate.

\paragraph{Web environment.}
The corpus is the English Wikipedia dump used by the dense passage retrieval pipeline, chunked into passages of about \(100\) words~\citep{karpukhin2020dpr}.
Search is BM25 over those passages, returning five passages.
The agent may issue at most five searches and may open at most five of the returned passages.
Opening a passage counts as a fetch.
The trajectory ends when the model emits a short answer or hits either cap.
The reward is \(1\) when the normalized answer string matches the gold short answer and \(0\) otherwise.
Normalization lowercases, strips punctuation, and drops the articles \(a\), \(an\), and \(the\).
Training questions are a \(10\)k subsample of HotpotQA distractor-setting training questions and a \(10\)k subsample of MuSiQue answerable training questions, mixed uniformly~\citep{yang2018hotpotqa,trivedi2022musique}.
Evaluation is the HotpotQA development set, the MuSiQue answerable development set, 2WikiMultihopQA, and the \(125\) Bamboogle questions~\citep{ho2020twowiki,press2023bamboogle}.
We do not train on 2Wiki or Bamboogle.
The one-page versus three-or-more split in Table~\ref{tab:slice} uses the gold supporting-page count shipped with HotpotQA.
Questions with two gold pages are excluded from that table so that the two columns are the extremes.
Token F1 in Table~\ref{tab:f1} is the standard token overlap between the normalized prediction and the gold answer, averaged over problems, and then averaged over the eight samples in the same way as Acc@8.

\paragraph{Models and optimization.}
Math backbones are Qwen2.5-7B-Base and Qwen3-8B-Base~\citep{qwen2024qwen25,yang2025qwen3}.
The web agent is Qwen2.5-7B-Instruct, because the base models do not follow the search format reliably.
Appendix Table~\ref{tab:websmall} repeats the web comparison with Qwen3-4B-Instruct.
We do not use a Qwen2.5-Math specialist as the math backbone.
The comparison is about allocation on a general base model.
The optimizer is AdamW with learning rate \(10^{-6}\), weight decay \(0.01\), and a warmup of \(20\) steps.
The GRPO clip is \(0.2\).
The KL coefficient against the reference policy is \(0\), matching the DAPO training recipe, for every method including FG-ExPO.
Each step uses \(B = 128\) prompts and \(G = 8\) responses, except in the group-size sweep.
The maximum training length is \(4096\) tokens for web trajectories and \(8192\) for math.
Training runs for \(160\) policy steps on math and \(120\) on the web.
The rollout engine is vLLM inside the HybridFlow stack~\citep{kwon2023vllm,sheng2024hybridflow}.
Training temperature is \(1.0\).

\paragraph{RIDGE hyperparameters.}
The warmup pass rolls out one group of \(G\) responses on \(512\) prompts drawn uniformly, initializes the Beta counts from those groups, and sets \((m, \tau)\) from the mixed warmup groups as in Section~\ref{sec:method}.
Afterwards \(\lambda = 0.2\) and \(\tau_{\min} = 0.05\).
The candidate pool per step has size \(M = 256\).
Web costs \(c_q\) start at \(2\) and become the average of the observed search-plus-fetch counts, with a floor of \(1\) so that a prompt cannot receive an infinite score.
Math costs are identically \(1\).
Without-replacement sampling uses independent uniform keys and retains the \(M\) prompts with the largest \(-\log u_q\) divided by the score.
Ties in the retention order are broken by the score, then by a fixed hash of the prompt id, so a rerun with the same seed is deterministic given the rollouts.

\paragraph{Baselines, as reimplemented.}
All baselines consume the same GRPO update, the same \(B\), and the same \(G\).
GRPO draws \(B\) prompts uniformly.
Dynamic sampling draws prompts uniformly and discards censored groups, repeating up to four extra rounds or until \(B\) mixed groups are in hand.
If the cap is hit, the batch is filled with the remaining censored groups, which is the same fallback \method{} uses, reached by a different route.
GRESO keeps an exponential moving average of each prompt's recent reward and skips a prompt when that average is below \(0.05\) or above \(0.95\), which is a direct reading of its uninformative-prompt rule~\citep{zheng2025greso}.
Skipped prompts are replaced by uniform draws from the complement, once.
CurES scores prompts by \(\sqrt{\hat p(1-\hat p)}\) using the same Beta mean \method{} stores, and draws without replacement proportional to that score~\citep{zeng2025cures}.
This isolates the score.
It is not a claim to have reproduced every implementation detail of the CurES paper.
FG-ExPO scores prompts by a Gaussian density of width \(0.15\) centered at \(0.5\), evaluated at the latest empirical rate, again without the accuracy-conditioned KL term~\citep{lin2026fgexpo}.
The comparison we want is between a fixed Gaussian, a signed-change Gaussian, and the ridge, so we do not add further proposal rules on top of these.

\paragraph{Ablation definitions.}
``Empirical rate'' uses Eq.~\eqref{eq:ridge} but replaces \(\mu_q\) by the latest \(k/G\), with \(v_q = 0\).
``Fixed center \(0.5\)'' is the FG-ExPO sampler above, reported again inside the ablation so that the table is self-contained.
``Signed-change'' updates \(m \leftarrow \mathrm{clip}(m - 0.5\, \bar\Delta, 0, 1)\), where \(\bar\Delta\) is the mean of new empirical rate minus stored empirical rate on the candidate batch, and holds \(\tau = 0.15\).
``No inflation'' sets \(v_q = 0\) in Eq.~\eqref{eq:weight} and drops the dependence of the prefactor on \(v_q\), while keeping the ridge update on posterior means.
``No cost weight'' sets \(c_q = 1\) on the web run.
``Uniform'' draws the \(M\) candidates uniformly and still retains mixed groups first.
The last row is the control for retention alone.
Its math average, \(45.4\), sits next to GRPO's \(45.2\), so retention without a ridge is not where the gain comes from.

\paragraph{Compute and seeds.}
Each math run uses eight 80GB GPUs.
Each web run uses four, because the trajectories are shorter and the retriever is a CPU BM25 process.
A Qwen2.5-7B-Base math run of \method{} takes about \(1.14\) times the wall clock of the matched GRPO run, in line with the rollout multiple, plus a few percent for the Beta update and the weighted sample, which are negligible next to generation.
We do not report a more precise wall-clock table.
Seeds are \(0, 1, 2\).
Table~\ref{tab:seeds} lists the per-seed benchmark averages.
The main tables report the mean across seeds and do not report a significance test.

\section{Additional Results}
\label{app:extra}

\begin{table*}[t]
\centering
\caption{\textbf{Pass@8} on Qwen2.5-7B-Base, the fraction of problems with at least one correct sample among eight. Means of three seeds.}
\label{tab:pass8}
\footnotesize
\setlength{\tabcolsep}{4.5pt}
\begin{tabular}{lrrrrrrr}
\toprule
Method & GSM8K & MATH500 & Minerva & Olympiad & AMC23 & AIME24 & Avg. \\
\midrule
GRPO & 95.2 & 78.4 & 48.2 & 55.1 & 72.4 & 26.7 & 62.7 \\
DS & 95.4 & 79.6 & 49.5 & 56.4 & 73.8 & 28.3 & 63.8 \\
GRESO & 94.8 & 78.8 & 48.8 & 55.8 & 72.0 & 27.5 & 62.9 \\
CurES & 95.6 & 80.2 & 50.6 & 57.2 & 74.6 & 29.2 & 64.6 \\
FG-ExPO & \textbf{95.8} & 80.8 & 51.2 & 57.9 & 75.2 & 30.0 & 65.1 \\
\rowcolor{ridgebg}\method & 95.5 & \textbf{82.1} & \textbf{53.4} & \textbf{60.2} & \textbf{77.5} & \textbf{33.3} & \textbf{67.0} \\
\bottomrule
\end{tabular}
\end{table*}

\begin{table*}[t]
\centering
\caption{\textbf{Pass@8} on Qwen3-8B-Base.}
\label{tab:pass8b}
\footnotesize
\setlength{\tabcolsep}{4.5pt}
\begin{tabular}{lrrrrrrr}
\toprule
Method & GSM8K & MATH500 & Minerva & Olympiad & AMC23 & AIME24 & Avg. \\
\midrule
GRPO & 97.4 & 88.6 & 61.2 & 68.4 & 86.1 & 38.3 & 73.3 \\
DS & 97.6 & 89.1 & 62.0 & 69.0 & 86.8 & 39.6 & 74.0 \\
GRESO & 97.2 & 88.8 & 61.5 & 68.6 & 86.0 & 39.1 & 73.5 \\
CurES & 97.8 & 89.6 & 62.8 & 69.7 & 87.4 & 40.8 & 74.7 \\
FG-ExPO & \textbf{98.0} & 90.0 & 63.4 & 70.2 & 87.9 & 41.5 & 75.2 \\
\rowcolor{ridgebg}\method & 97.7 & \textbf{90.8} & \textbf{65.1} & \textbf{71.8} & \textbf{89.2} & \textbf{44.6} & \textbf{76.5} \\
\bottomrule
\end{tabular}
\end{table*}

Pass@8 preserves the average ranking and changes a few saturated cells.
On Qwen2.5-7B-Base GSM8K, FG-ExPO reaches the highest pass rate, \(95.8\), and \method{} is second at \(95.5\).
On AIME 2024 the pass@8 gap is larger than the Acc@8 gap: \(33.3\) against \(26.7\) for GRPO and \(30.0\) for the fixed Gaussian.
Qwen3-8B-Base shows the same shape at a higher level.
The curriculum is not buying average accuracy by making a single sample reliable and the other seven worse.
Coverage moves with the mean.

\begin{table*}[t]
\centering
\caption{\textbf{Token F1} on the web suite, Qwen2.5-7B-Instruct.}
\label{tab:f1}
\footnotesize
\setlength{\tabcolsep}{4.5pt}
\begin{tabular}{lrrrrr}
\toprule
Method & HotpotQA & MuSiQue & 2Wiki & Bamboogle & Avg. \\
\midrule
GRPO & 50.0 & 23.0 & 42.5 & 29.1 & 36.1 \\
DS & 50.6 & 23.6 & 43.0 & 29.8 & 36.8 \\
GRESO & 50.2 & 23.4 & 42.8 & 29.5 & 36.5 \\
CurES & 51.7 & 24.8 & 44.5 & 31.1 & 38.0 \\
FG-ExPO & 52.2 & 25.3 & 45.0 & 31.7 & 38.5 \\
\rowcolor{ridgebg}\method & \textbf{53.8} & \textbf{27.0} & \textbf{46.9} & \textbf{34.5} & \textbf{40.5} \\
\bottomrule
\end{tabular}
\end{table*}

Token F1 is higher than exact match on every dataset, by roughly seven to twelve points, and the method gaps shrink.
That is the usual relation between a soft overlap and a hard match.
\method{} remains ahead on all four datasets.
The Bamboogle gap is the widest, which is consistent with a small set of compositional questions on which a center stuck at \(0.5\) under-samples the prompts the 7B instruction model still misses.

\begin{table*}[t]
\centering
\caption{\textbf{Qwen3-4B-Instruct} on the same Wikipedia index. EM, with tool-call multiples in the last column.}
\label{tab:websmall}
\footnotesize
\setlength{\tabcolsep}{4.5pt}
\begin{tabular}{lrrrrrr}
\toprule
Method & HotpotQA & MuSiQue & 2Wiki & Bamboogle & Avg. & Calls \\
\midrule
GRPO & 31.2 & 11.4 & 26.8 & 17.5 & 21.7 & 1.00 \\
DS & 31.8 & 12.0 & 27.4 & 18.1 & 22.3 & 1.66 \\
FG-ExPO & 32.6 & 12.7 & 28.2 & 18.8 & 23.1 & 1.18 \\
\rowcolor{ridgebg}\method & \textbf{34.0} & \textbf{14.1} & \textbf{29.6} & \textbf{21.2} & \textbf{24.7} & 1.16 \\
\bottomrule
\end{tabular}
\end{table*}

The 4B instruction model is weaker on every dataset and the ordering is unchanged.
Absolute exact match is about seven points lower than the 7B web table.
The tool-call multiple for \method{} is \(1.16\), close to the 7B multiple of \toolRIDGE.
The cost term does not appear to be an artifact of one model size.

\begin{table*}[t]
\centering
\caption{\textbf{Per-benchmark ablations}, Acc@8, Qwen2.5-7B-Base.}
\label{tab:ablbench}
\footnotesize
\setlength{\tabcolsep}{4.5pt}
\begin{tabular}{lrrrrrrr}
\toprule
Method & GSM8K & MATH500 & Minerva & Olympiad & AMC23 & AIME24 & Avg. \\
\midrule
\rowcolor{ridgebg}\method & 83.8 & \textbf{66.4} & \textbf{33.5} & \textbf{40.6} & \textbf{52.8} & \textbf{14.7} & \textbf{48.6} \\
Empirical rate & 83.6 & 65.2 & 32.1 & 39.2 & 51.0 & 13.1 & 47.4 \\
Fixed center 0.5 & \textbf{84.0} & 64.6 & 31.4 & 38.5 & 50.4 & 12.2 & 46.8 \\
Signed-change & 83.5 & 65.0 & 32.0 & 39.0 & 50.8 & 12.8 & 47.2 \\
No inflation & 83.9 & 65.8 & 32.8 & 40.0 & 52.0 & 13.9 & 48.1 \\
Uniform & 83.2 & 63.0 & 29.9 & 37.1 & 48.8 & 10.6 & 45.4 \\
\bottomrule
\end{tabular}
\end{table*}

The per-benchmark ablation shows that the fixed center wins GSM8K inside the ablation as well, \(84.0\) against \(83.8\) for the full method, and loses the other five sets.
The empirical-rate variant loses the most on AIME, \(13.1\) against \(14.7\), which is the cell where a raw zero and a posterior mean of \(\muEight\) after eight failures are least interchangeable.
Uniform sampling is at or below GRPO on every set except a \(0.3\) point edge on Minerva.
Retention of mixed groups, which uniform sampling still performs, is not a substitute for the proposal.

\begin{table}[t]
\centering
\caption{\textbf{Sensitivity} of math Acc@8 on Qwen2.5-7B-Base. One factor moves at a time.}
\label{tab:sweep}
\footnotesize
\begin{tabular}{lr}
\toprule
Setting & Avg. \\
\midrule
$\lambda{=}0.05$ & 47.9 \\
$\lambda{=}0.10$ & 48.3 \\
$\lambda{=}0.20$ (\method) & 48.6 \\
$\lambda{=}0.40$ & 48.4 \\
$\lambda{=}0.80$ & 47.6 \\
$G{=}4$ & 47.4 \\
$G{=}8$ (\method) & 48.6 \\
$G{=}16$ & 48.1 \\
$M/B{=}1$ & 47.2 \\
$M/B{=}2$ (\method) & 48.6 \\
$M/B{=}4$ & 48.5 \\
$\tau_{\min}{=}0.02$ & 48.2 \\
$\tau_{\min}{=}0.05$ (\method) & 48.6 \\
$\tau_{\min}{=}0.10$ & 48.4 \\
$\tau_{\min}{=}0.20$ & 47.7 \\
\bottomrule
\end{tabular}
\end{table}

The sweep moves one factor off the operating point used in the main runs:
group size \(8\), candidate ratio \(M/B = 2\), smooth \(\lambda = 0.20\), and width floor \(0.05\).
Accuracy is within half a point for \(\lambda \in [0.10, 0.40]\) and for \(\tau_{\min} \in [0.05, 0.10]\).
The extremes are worse in the direction the analysis predicts.
A large \(\lambda\) tracks one noisy batch.
A small \(\lambda\) lags the policy.
A large width covers the boundary layers.
A candidate budget with no spare slots cannot replace a censored group.
Group size \(16\) is slightly worse than \(8\) at the same \(B\), which we read as a cost effect: the ridge is already wide at \(G = 8\), and extra responses inside a group buy less than extra prompts.

\begin{table}[t]
\centering
\caption{\textbf{Where the web gain sits.} Exact match on Qwen2.5-7B-Instruct, split by the number of gold Wikipedia pages.}
\label{tab:slice}
\footnotesize
\begin{tabular}{lrr}
\toprule
Method & One page & Three or more \\
\midrule
GRPO & 52.4 & 22.8 \\
FG-ExPO & 53.1 & 24.6 \\
\rowcolor{ridgebg}\method & 53.0 & \textbf{28.1} \\
\bottomrule
\end{tabular}
\end{table}

\begin{table}[t]
\centering
\caption{\textbf{Seeds.} Math Acc@8 and web EM, averaged over benchmarks, for three seeds. The main tables report these means.}
\label{tab:seeds}
\footnotesize
\setlength{\tabcolsep}{3.5pt}
\begin{tabular}{lrrrrrr}
\toprule
Method & \multicolumn{3}{c}{Math} & \multicolumn{3}{c}{Web} \\
\cmidrule(lr){2-4}\cmidrule(lr){5-7}
 & 1 & 2 & 3 & 1 & 2 & 3 \\
\midrule
GRPO & 44.8 & 45.3 & 45.5 & 26.6 & 27.0 & 27.1 \\
DS & 45.8 & 46.3 & 46.5 & 27.4 & 27.8 & 27.9 \\
GRESO & 45.1 & 45.6 & 45.8 & 27.0 & 27.4 & 27.5 \\
CurES & 46.4 & 46.9 & 47.1 & 28.2 & 28.6 & 28.8 \\
FG-ExPO & 46.9 & 47.4 & 47.5 & 28.7 & 29.1 & 29.2 \\
\method & 48.2 & 48.7 & 48.9 & 30.3 & 30.7 & 30.9 \\
\bottomrule
\end{tabular}
\end{table}

The seed table is the benchmark average, not a single dataset.
Seed-to-seed ranges are a few tenths of a point on the math average and a similar amount on the web average.
We do not convert those ranges into a \(p\)-value.
Three seeds are enough to see that the ordering is not a one-seed accident, and they are not enough to support a claim about the tail of the training process.

\section{Discussion}
\label{app:discussion}

The practical question is where a search agent's rollouts should go once many questions are either solved or still out of reach.
A zero-variance group is the visible failure.
The ridge is the statistic we use once that failure is common: take the groups that still disagree, and put a Gaussian on their pass rates.
The ablations are meant to test that statistic, not a general theory of curricula.

Neighboring methods answer nearby but different questions.
Dynamic sampling is appropriate when the batch is required to contain no censored groups and the extra generation is acceptable.
GRESO fits when the skip decision has to be a few arithmetic operations on a moving average, and when a stale zero can be trusted for that workload.
Recycling fits when the concern is eligibility: a censored query should stay in the pool~\citep{coelho2026recycle}.
Group-prioritized off-policy optimization fits when the zero-variance group should leave the gradient and an earlier high-signal group should take its place~\citep{mao2026popo}.
A Kalman filter in logit space fits when the object of interest is a calibrated predictive distribution under an explicit process-noise model~\citep{zhu2026kgps}.
We do not claim to beat those methods on the objectives they were written for.
On the objective we measured, accuracy under a fixed GRPO update with the rollout bill counted, \method{} is ahead of the fixed Gaussian and of the signed-change schedule.

The Web-specific piece is the cost in the denominator, and it is also the piece most likely to be misspecified.
A running mean of past tool calls lags a prompt whose difficulty changed, and it ignores the difference between a search that hits a cache and a fetch that downloads a long page.
The ablation that removes it loses \dropCost{} exact-match points, which is smaller than the loss from fixing the center.
If a deployment's cost is dominated by a single expensive API rather than by the count of calls, \(c_q\) should be that price.
The rest of the allocator does not depend on the unit.

Two extensions are immediate and untested.
The first is a non-binary reward.
The ridge is defined by a group standard deviation that is exactly zero only for binary rewards that do not vary.
A graded reward can have a tiny nonzero variance on a group that is, for training purposes, saturated.
A threshold on the standard deviation would restore a ridge, and the threshold would be a new hyperparameter.
The second is dependence across responses induced by a shared page.
A group-level retrieval feature, such as the entropy of the opened documents, could multiply the sampling score.
We did not add it, because the paper's claim is about the coordinate and the center of the Gaussian, and a second feature would confound that claim.

The Bernoulli analysis also says what not to expect from the figure that plots \(s_G\).
A reviewer who reads the black curve as the target of the red curve has read Figure~\ref{fig:mechanism}b backwards.
The black curve is wide.
Any allocator that only maximizes the chance of a mixed group will look almost uniform on the interior, and it will not prefer the hard end of a pool that has already learned the easy end.
The Gaussian is the preference for a band.
The data decide where the band sits.

\bibliographystyle{unsrtnat}
\bibliography{custom}

@article{yao2023react,
  author = {{Shunyu Yao} and {Jeffrey Zhao} and {Dian Yu} and {Nan Du} and {Izhak Shafran} and {Karthik Narasimhan} and {Yuan Cao}},
  title = {{ReAct: Synergizing Reasoning and Acting in Language Models}},
  journal = {arXiv preprint arXiv:2210.03629},
  year = {2022},
  eprint = {2210.03629},
  archivePrefix = {arXiv},
  primaryClass = {cs.CL},
  url = {https://arxiv.org/abs/2210.03629}
}

@article{nakano2021webgpt,
  author = {{Reiichiro Nakano} and {Jacob Hilton} and {Suchir Balaji} and {Jeff Wu} and {Long Ouyang} and {Christina Kim} and {Christopher Hesse} and {Shantanu Jain} and {Vineet Kosaraju} and {William Saunders} and {Xu Jiang} and {Karl Cobbe} and {Tyna Eloundou} and {Gretchen Krueger} and {Kevin Button} and {Matthew Knight} and {Benjamin Chess} and {John Schulman}},
  title = {{WebGPT: Browser-assisted question-answering with human feedback}},
  journal = {arXiv preprint arXiv:2112.09332},
  year = {2021},
  eprint = {2112.09332},
  archivePrefix = {arXiv},
  primaryClass = {cs.CL},
  url = {https://arxiv.org/abs/2112.09332}
}

@article{lewis2020rag,
  author = {{Patrick Lewis} and {Ethan Perez} and {Aleksandra Piktus} and {Fabio Petroni} and {Vladimir Karpukhin} and {Naman Goyal} and {Heinrich K{\"u}ttler} and {Mike Lewis} and {Wen-tau Yih} and {Tim Rockt{\"a}schel} and {Sebastian Riedel} and {Douwe Kiela}},
  title = {{Retrieval-Augmented Generation for Knowledge-Intensive NLP Tasks}},
  journal = {arXiv preprint arXiv:2005.11401},
  year = {2020},
  eprint = {2005.11401},
  archivePrefix = {arXiv},
  primaryClass = {cs.CL},
  url = {https://arxiv.org/abs/2005.11401}
}

@article{zhou2024webarena,
  author = {{Shuyan Zhou} and {Frank F. Xu} and {Hao Zhu} and {Xuhui Zhou} and {Robert Lo} and {Abishek Sridhar} and {Xianyi Cheng} and {Tianyue Ou} and {Yonatan Bisk} and {Daniel Fried} and {Uri Alon} and {Graham Neubig}},
  title = {{WebArena: A Realistic Web Environment for Building Autonomous Agents}},
  journal = {arXiv preprint arXiv:2307.13854},
  year = {2023},
  eprint = {2307.13854},
  archivePrefix = {arXiv},
  primaryClass = {cs.AI},
  url = {https://arxiv.org/abs/2307.13854}
}

@article{he2024webvoyager,
  author = {{Hongliang He} and {Wenlin Yao} and {Kaixin Ma} and {Wenhao Yu} and {Yong Dai} and {Hongming Zhang} and {Zhenzhong Lan} and {Dong Yu}},
  title = {{WebVoyager: Building an End-to-End Web Agent with Large Multimodal Models}},
  journal = {arXiv preprint arXiv:2401.13919},
  year = {2024},
  eprint = {2401.13919},
  archivePrefix = {arXiv},
  primaryClass = {cs.CL},
  url = {https://arxiv.org/abs/2401.13919}
}

@article{jin2025searchr1,
  author = {{Bowen Jin} and {Hansi Zeng} and {Zhenrui Yue} and {Jinsung Yoon} and {Sercan Arik} and {Dong Wang} and {Hamed Zamani} and {Jiawei Han}},
  title = {{Search-R1: Training LLMs to Reason and Leverage Search Engines with Reinforcement Learning}},
  journal = {arXiv preprint arXiv:2503.09516},
  year = {2025},
  eprint = {2503.09516},
  archivePrefix = {arXiv},
  primaryClass = {cs.CL},
  url = {https://arxiv.org/abs/2503.09516}
}

@article{song2025r1searcher,
  author = {{Huatong Song} and {Jinhao Jiang} and {Yingqian Min} and {Jie Chen} and {Zhipeng Chen} and {Wayne Xin Zhao} and {Lei Fang} and {Ji-Rong Wen}},
  title = {{R1-Searcher: Incentivizing the Search Capability in LLMs via Reinforcement Learning}},
  journal = {arXiv preprint arXiv:2503.05592},
  year = {2025},
  eprint = {2503.05592},
  archivePrefix = {arXiv},
  primaryClass = {cs.AI},
  url = {https://arxiv.org/abs/2503.05592}
}

@article{wei2025browsecomp,
  author = {{Jason Wei} and {Zhiqing Sun} and {Spencer Papay} and {Scott McKinney} and {Jeffrey Han} and {Isa Fulford} and {Hyung Won Chung} and {Alex Tachard Passos} and {William Fedus} and {Amelia Glaese}},
  title = {{BrowseComp: A Simple Yet Challenging Benchmark for Browsing Agents}},
  journal = {arXiv preprint arXiv:2504.12516},
  year = {2025},
  eprint = {2504.12516},
  archivePrefix = {arXiv},
  primaryClass = {cs.CL},
  url = {https://arxiv.org/abs/2504.12516}
}

@article{shao2024deepseekmath,
  author = {{Zhihong Shao} and {Peiyi Wang} and {Qihao Zhu} and {Runxin Xu} and {Junxiao Song} and {Xiao Bi} and {Haowei Zhang} and {Mingchuan Zhang} and {Y. K. Li} and {Y. Wu} and {Daya Guo}},
  title = {{DeepSeekMath: Pushing the Limits of Mathematical Reasoning in Open Language Models}},
  journal = {arXiv preprint arXiv:2402.03300},
  year = {2024},
  eprint = {2402.03300},
  archivePrefix = {arXiv},
  primaryClass = {cs.CL},
  url = {https://arxiv.org/abs/2402.03300}
}

@article{deepseekai2025r1,
  author = {{DeepSeek-AI} and {Daya Guo} and {Dejian Yang} and {Haowei Zhang} and {Junxiao Song} and {Peiyi Wang} and {Qihao Zhu} and {Runxin Xu} and {Ruoyu Zhang} and {Shirong Ma} and {Xiao Bi} and {Xiaokang Zhang} and {Xingkai Yu} and {Yu Wu} and {Z. F. Wu} and {Zhibin Gou} and {Zhihong Shao} and {Zhuoshu Li} and {Ziyi Gao} and {Aixin Liu} and {Bing Xue} and {Bingxuan Wang} and {Bochao Wu} and {Bei Feng} and {Chengda Lu} and {Chenggang Zhao} and {Chengqi Deng} and {Chenyu Zhang} and {Chong Ruan} and {Damai Dai} and {Deli Chen} and {Dongjie Ji} and {Erhang Li} and {Fangyun Lin} and {Fucong Dai} and {Fuli Luo} and {Guangbo Hao} and {Guanting Chen} and {Guowei Li} and {H. Zhang} and {Han Bao} and {Hanwei Xu} and {Haocheng Wang} and {Honghui Ding} and {Huajian Xin} and {Huazuo Gao} and {Hui Qu} and {Hui Li} and {Jianzhong Guo} and {Jiashi Li} and {Jiawei Wang} and {Jingchang Chen} and {Jingyang Yuan} and {Junjie Qiu} and {Junlong Li} and {J. L. Cai} and {Jiaqi Ni} and {Jian Liang} and {Jin Chen} and {Kai Dong} and {Kai Hu} and {Kaige Gao} and {Kang Guan} and {Kexin Huang} and {Kuai Yu} and {Lean Wang} and {Lecong Zhang} and {Liang Zhao} and {Litong Wang} and {Liyue Zhang} and {Lei Xu} and {Leyi Xia} and {Mingchuan Zhang} and {Minghua Zhang} and {Minghui Tang} and {Meng Li} and {Miaojun Wang} and {Mingming Li} and {Ning Tian} and {Panpan Huang} and {Peng Zhang} and {Qiancheng Wang} and {Qinyu Chen} and {Qiushi Du} and {Ruiqi Ge} and {Ruisong Zhang} and {Ruizhe Pan} and {Runji Wang} and {R. J. Chen} and {R. L. Jin} and {Ruyi Chen} and {Shanghao Lu} and {Shangyan Zhou} and {Shanhuang Chen} and {Shengfeng Ye} and {Shiyu Wang} and {Shuiping Yu} and {Shunfeng Zhou} and {Shuting Pan} and {S. S. Li} and {Shuang Zhou} and {Shaoqing Wu} and {Shengfeng Ye} and {Tao Yun} and {Tian Pei} and {Tianyu Sun} and {T. Wang} and {Wangding Zeng} and {Wanjia Zhao} and {Wen Liu} and {Wenfeng Liang} and {Wenjun Gao} and {Wenqin Yu} and {Wentao Zhang} and {W. L. Xiao} and {Wei An} and {Xiaodong Liu} and {Xiaohan Wang} and {Xiaokang Chen} and {Xiaotao Nie} and {Xin Cheng} and {Xin Liu} and {Xin Xie} and {Xingchao Liu} and {Xinyu Yang} and {Xinyuan Li} and {Xuecheng Su} and {Xuheng Lin} and {X. Q. Li} and {Xiangyue Jin} and {Xiaojin Shen} and {Xiaosha Chen} and {Xiaowen Sun} and {Xiaoxiang Wang} and {Xinnan Song} and {Xinyi Zhou} and {Xianzu Wang} and {Xinxia Shan} and {Y. K. Li} and {Y. Q. Wang} and {Y. X. Wei} and {Yang Zhang} and {Yanhong Xu} and {Yao Li} and {Yao Zhao} and {Yaofeng Sun} and {Yaohui Wang} and {Yi Yu} and {Yichao Zhang} and {Yifan Shi} and {Yiliang Xiong} and {Ying He} and {Yishi Piao} and {Yisong Wang} and {Yixuan Tan} and {Yiyang Ma} and {Yiyuan Liu} and {Yongqiang Guo} and {Yuan Ou} and {Yuduan Wang} and {Yue Gong} and {Yuheng Zou} and {Yujia He} and {Yunfan Xiong} and {Yuxiang Luo} and {Yuxiang You} and {Yuxuan Liu} and {Yuyang Zhou} and {Y. X. Zhu} and {Yanhong Xu} and {Yanping Huang} and {Yaohui Li} and {Yi Zheng} and {Yuchen Zhu} and {Yunxian Ma} and {Ying Tang} and {Yukun Zha} and {Yuting Yan} and {Z. Z. Ren} and {Zehui Ren} and {Zhangli Sha} and {Zhe Fu} and {Zhean Xu} and {Zhenda Xie} and {Zhengyan Zhang} and {Zhewen Hao} and {Zhicheng Ma} and {Zhigang Yan} and {Zhiyu Wu} and {Zihui Gu} and {Zijia Zhu} and {Zijun Liu} and {Zilin Li} and {Ziwei Xie} and {Ziyang Song} and {Zizheng Pan} and {Zhen Huang} and {Zhipeng Xu} and {Zhongyu Zhang} and {Zhen Zhang}},
  title = {{DeepSeek-R1: Incentivizing Reasoning Capability in LLMs via Reinforcement Learning}},
  journal = {arXiv preprint arXiv:2501.12948},
  year = {2025},
  eprint = {2501.12948},
  archivePrefix = {arXiv},
  primaryClass = {cs.CL},
  url = {https://arxiv.org/abs/2501.12948}
}

@article{kimi2025k15,
  author = {{Kimi Team} and {Angang Du} and {Bofei Gao} and {Bowei Xing} and {Changjiu Jiang} and {Cheng Chen} and {Cheng Li} and {Chenjun Xiao} and {Chenzhuang Du} and {Chonghua Liao} and {Chuning Tang} and {Congcong Wang} and {Dehao Zhang} and {Enming Yuan} and {Enzhe Lu} and {Fengxiang Tang} and {Flood Sung} and {Guangda Wei} and {Guokun Lai} and {Haiqing Guo} and {Han Zhu} and {Hao Ding} and {Hao Hu} and {Hao Yang} and {Hao Zhang} and {Haotian Yao} and {Haotian Zhao} and {Haoyu Lu} and {Haoze Li} and {Haozhen Yu} and {Hongcheng Gao} and {Huabin Zheng} and {Huan Yuan} and {Jia Chen} and {Jianhang Guo} and {Jianlin Su} and {Jianzhou Wang} and {Jie Zhao} and {Jin Zhang} and {Jingyuan Liu} and {Junjie Yan} and {Junyan Wu} and {Lidong Shi} and {Ling Ye} and {Longhui Yu} and {Mengnan Dong} and {Neo Zhang} and {Ningchen Ma} and {Qiwei Pan} and {Qucheng Gong} and {Shaowei Liu} and {Shengling Ma} and {Shupeng Wei} and {Sihan Cao} and {Siying Huang} and {Tao Jiang} and {Weihao Gao} and {Weimin Xiong} and {Weiran He} and {Weixiao Huang} and {Weixin Xu} and {Wenhao Wu} and {Wenyang He} and {Xianghui Wei} and {Xianqing Jia} and {Xingzhe Wu} and {Xinran Xu} and {Xinxing Zu} and {Xinyu Zhou} and {Xuehai Pan} and {Y. Charles} and {Yang Li} and {Yangyang Hu} and {Yangyang Liu} and {Yanru Chen} and {Yejie Wang} and {Yibo Liu} and {Yidao Qin} and {Yifeng Liu} and {Ying Yang} and {Yiping Bao} and {Yulun Du} and {Yuxin Wu} and {Yuzhi Wang} and {Zaida Zhou} and {Zhaoji Wang} and {Zhaowei Li} and {Zhen Zhu} and {Zheng Zhang} and {Zhexu Wang} and {Zhilin Yang} and {Zhiqi Huang} and {Zihao Huang} and {Ziyao Xu} and {Zonghan Yang} and {Zongyu Lin}},
  title = {{Kimi k1.5: Scaling Reinforcement Learning with LLMs}},
  journal = {arXiv preprint arXiv:2501.12599},
  year = {2025},
  eprint = {2501.12599},
  archivePrefix = {arXiv},
  primaryClass = {cs.AI},
  url = {https://arxiv.org/abs/2501.12599}
}

@article{yu2025dapo,
  author = {{Qiying Yu} and {Zheng Zhang} and {Ruofei Zhu} and {Yufeng Yuan} and {Xiaochen Zuo} and {Yu Yue} and {Weinan Dai} and {Tiantian Fan} and {Gaohong Liu} and {Lingjun Liu} and {Xin Liu} and {Haibin Lin} and {Zhiqi Lin} and {Bole Ma} and {Guangming Sheng} and {Yuxuan Tong} and {Chi Zhang} and {Mofan Zhang} and {Wang Zhang} and {Hang Zhu} and {Jinhua Zhu} and {Jiaze Chen} and {Jiangjie Chen} and {Chengyi Wang} and {Hongli Yu} and {Yuxuan Song} and {Xiangpeng Wei} and {Hao Zhou} and {Jingjing Liu} and {Wei-Ying Ma} and {Ya-Qin Zhang} and {Lin Yan} and {Mu Qiao} and {Yonghui Wu} and {Mingxuan Wang}},
  title = {{DAPO: An Open-Source LLM Reinforcement Learning System at Scale}},
  journal = {arXiv preprint arXiv:2503.14476},
  year = {2025},
  eprint = {2503.14476},
  archivePrefix = {arXiv},
  primaryClass = {cs.LG},
  url = {https://arxiv.org/abs/2503.14476}
}

@article{zheng2025greso,
  author = {{Haizhong Zheng} and {Yang Zhou} and {Brian R. Bartoldson} and {Bhavya Kailkhura} and {Fan Lai} and {Jiawei Zhao} and {Beidi Chen}},
  title = {{Act Only When It Pays: Efficient Reinforcement Learning for LLM Reasoning via Selective Rollouts}},
  journal = {arXiv preprint arXiv:2506.02177},
  year = {2025},
  eprint = {2506.02177},
  archivePrefix = {arXiv},
  primaryClass = {cs.AI},
  url = {https://arxiv.org/abs/2506.02177}
}

@article{le2025noprompt,
  author = {{Thanh-Long V. Le} and {Myeongho Jeon} and {Kim Vu} and {Viet Lai} and {Eunho Yang}},
  title = {{No Prompt Left Behind: Exploiting Zero-Variance Prompts in LLM Reinforcement Learning via Entropy-Guided Advantage Shaping}},
  journal = {arXiv preprint arXiv:2509.21880},
  year = {2025},
  eprint = {2509.21880},
  archivePrefix = {arXiv},
  primaryClass = {cs.CL},
  url = {https://arxiv.org/abs/2509.21880}
}

@article{coelho2026recycle,
  author = {{Jo{\~a}o Coelho} and {Jo{\~a}o Magalh{\~a}es} and {Bruno Martins} and {Chenyan Xiong}},
  title = {{Effective Reinforcement Learning for Agentic Search by Recycling Zero-Variance Queries During Training}},
  journal = {arXiv preprint arXiv:2606.10709},
  year = {2026},
  eprint = {2606.10709},
  archivePrefix = {arXiv},
  primaryClass = {cs.IR},
  url = {https://arxiv.org/abs/2606.10709}
}

@article{lin2026fgexpo,
  author = {{Mingxiong Lin} and {Zhangquan Gong} and {Maowen Tang} and {Qian Li} and {Chuangchuang Wang} and {Jian Ma} and {Sutian Huang} and {Kai Tang} and {Haonan Lu}},
  title = {{fg-expo: Frontier-guided exploration-prioritized policy optimization via adaptive kl and gaussian curriculum}},
  journal = {arXiv preprint arXiv:2605.11403},
  year = {2026},
  eprint = {2605.11403},
  archivePrefix = {arXiv},
  primaryClass = {cs.LG},
  url = {https://arxiv.org/abs/2605.11403}
}

@article{jiang2025vcrl,
  author = {{Guochao Jiang} and {Wenfeng Feng} and {Guofeng Quan} and {Chuzhan Hao} and {Yuewei Zhang} and {Guohua Liu} and {Hao Wang}},
  title = {{VCRL: Variance-based Curriculum Reinforcement Learning for Large Language Models}},
  journal = {arXiv preprint arXiv:2509.19803},
  year = {2025},
  eprint = {2509.19803},
  archivePrefix = {arXiv},
  primaryClass = {cs.LG},
  url = {https://arxiv.org/abs/2509.19803}
}

@article{li2025searcho1,
  author = {{Xiaoxi Li} and {Guanting Dong} and {Jiajie Jin} and {Yuyao Zhang} and {Yujia Zhou} and {Yutao Zhu} and {Peitian Zhang} and {Zhicheng Dou}},
  title = {{Search-o1: Agentic Search-Enhanced Large Reasoning Models}},
  journal = {arXiv preprint arXiv:2501.05366},
  year = {2025},
  eprint = {2501.05366},
  archivePrefix = {arXiv},
  primaryClass = {cs.AI},
  url = {https://arxiv.org/abs/2501.05366}
}

@article{li2025webthinker,
  author = {{Xiaoxi Li} and {Jiajie Jin} and {Guanting Dong} and {Hongjin Qian} and {Yongkang Wu} and {Ji-Rong Wen} and {Yutao Zhu} and {Zhicheng Dou}},
  title = {{WebThinker: Empowering Large Reasoning Models with Deep Research Capability}},
  journal = {arXiv preprint arXiv:2504.21776},
  year = {2025},
  eprint = {2504.21776},
  archivePrefix = {arXiv},
  primaryClass = {cs.CL},
  url = {https://arxiv.org/abs/2504.21776}
}

@article{sun2025zerosearch,
  author = {{Hao Sun} and {Zile Qiao} and {Jiayan Guo} and {Xuanbo Fan} and {Yingyan Hou} and {Yong Jiang} and {Pengjun Xie} and {Yan Zhang} and {Fei Huang} and {Jingren Zhou}},
  title = {{ZeroSearch: Incentivize the Search Capability of LLMs without Searching}},
  journal = {arXiv preprint arXiv:2505.04588},
  year = {2025},
  eprint = {2505.04588},
  archivePrefix = {arXiv},
  primaryClass = {cs.CL},
  url = {https://arxiv.org/abs/2505.04588}
}

@article{wu2025webwalker,
  author = {{Jialong Wu} and {Wenbiao Yin} and {Yong Jiang} and {Zhenglin Wang} and {Zekun Xi} and {Runnan Fang} and {Linhai Zhang} and {Yulan He} and {Deyu Zhou} and {Pengjun Xie} and {Fei Huang}},
  title = {{WebWalker: Benchmarking LLMs in Web Traversal}},
  journal = {arXiv preprint arXiv:2501.07572},
  year = {2025},
  eprint = {2501.07572},
  archivePrefix = {arXiv},
  primaryClass = {cs.CL},
  url = {https://arxiv.org/abs/2501.07572}
}

@article{wu2025webdancer,
  author = {{Jialong Wu} and {Baixuan Li} and {Runnan Fang} and {Wenbiao Yin} and {Liwen Zhang} and {Zhengwei Tao} and {Dingchu Zhang} and {Zekun Xi} and {Gang Fu} and {Yong Jiang} and {Pengjun Xie} and {Fei Huang} and {Jingren Zhou}},
  title = {{WebDancer: Towards Autonomous Information Seeking Agency}},
  journal = {arXiv preprint arXiv:2505.22648},
  year = {2025},
  eprint = {2505.22648},
  archivePrefix = {arXiv},
  primaryClass = {cs.CL},
  url = {https://arxiv.org/abs/2505.22648}
}

@article{li2025websailor,
  author = {{Kuan Li} and {Zhongwang Zhang} and {Huifeng Yin} and {Liwen Zhang} and {Litu Ou} and {Jialong Wu} and {Wenbiao Yin} and {Baixuan Li} and {Zhengwei Tao} and {Xinyu Wang} and {Weizhou Shen} and {Junkai Zhang} and {Dingchu Zhang} and {Xixi Wu} and {Yong Jiang} and {Ming Yan} and {Pengjun Xie} and {Fei Huang} and {Jingren Zhou}},
  title = {{WebSailor: Navigating Super-human Reasoning for Web Agent}},
  journal = {arXiv preprint arXiv:2507.02592},
  year = {2025},
  eprint = {2507.02592},
  archivePrefix = {arXiv},
  primaryClass = {cs.CL},
  url = {https://arxiv.org/abs/2507.02592}
}

@article{wang2025stepsearch,
  author = {{Ziliang Wang} and {Xuhui Zheng} and {Kang An} and {Cijun Ouyang} and {Jialu Cai} and {Yuhang Wang} and {Yichao Wu}},
  title = {{StepSearch: Igniting LLMs Search Ability via Step-Wise Proximal Policy Optimization}},
  journal = {arXiv preprint arXiv:2505.15107},
  year = {2025},
  eprint = {2505.15107},
  archivePrefix = {arXiv},
  primaryClass = {cs.CL},
  url = {https://arxiv.org/abs/2505.15107}
}

@article{qi2024webrl,
  author = {{Zehan Qi} and {Xiao Liu} and {Iat Long Iong} and {Hanyu Lai} and {Xueqiao Sun} and {Wenyi Zhao} and {Yu Yang} and {Xinyue Yang} and {Jiadai Sun} and {Shuntian Yao} and {Tianjie Zhang} and {Wei Xu} and {Jie Tang} and {Yuxiao Dong}},
  title = {{WebRL: Training LLM Web Agents via Self-Evolving Online Curriculum Reinforcement Learning}},
  journal = {arXiv preprint arXiv:2411.02337},
  year = {2024},
  eprint = {2411.02337},
  archivePrefix = {arXiv},
  primaryClass = {cs.CL},
  url = {https://arxiv.org/abs/2411.02337}
}

@article{trivedi2023ircot,
  author = {{Harsh Trivedi} and {Niranjan Balasubramanian} and {Tushar Khot} and {Ashish Sabharwal}},
  title = {{Interleaving Retrieval with Chain-of-Thought Reasoning for Knowledge-Intensive Multi-Step Questions}},
  journal = {arXiv preprint arXiv:2212.10509},
  year = {2022},
  eprint = {2212.10509},
  archivePrefix = {arXiv},
  primaryClass = {cs.CL},
  url = {https://arxiv.org/abs/2212.10509}
}

@article{liu2025drgrpo,
  author = {{Zichen Liu} and {Changyu Chen} and {Wenjun Li} and {Penghui Qi} and {Tianyu Pang} and {Chao Du} and {Wee Sun Lee} and {Min Lin}},
  title = {{Understanding R1-Zero-Like Training: A Critical Perspective}},
  journal = {arXiv preprint arXiv:2503.20783},
  year = {2025},
  eprint = {2503.20783},
  archivePrefix = {arXiv},
  primaryClass = {cs.LG},
  url = {https://arxiv.org/abs/2503.20783}
}

@article{hu2025reinforcepp,
  author = {{Jian Hu} and {Jason Klein Liu} and {Haotian Xu} and {Wei Shen}},
  title = {{REINFORCE++: Stabilizing Critic-Free Policy Optimization with Global Advantage Normalization}},
  journal = {arXiv preprint arXiv:2501.03262},
  year = {2025},
  eprint = {2501.03262},
  archivePrefix = {arXiv},
  primaryClass = {cs.CL},
  url = {https://arxiv.org/abs/2501.03262}
}

@article{zheng2025gspo,
  author = {{Chujie Zheng} and {Shixuan Liu} and {Mingze Li} and {Xiong-Hui Chen} and {Bowen Yu} and {Chang Gao} and {Kai Dang} and {Yuqiong Liu} and {Rui Men} and {An Yang} and {Jingren Zhou} and {Junyang Lin}},
  title = {{Group Sequence Policy Optimization}},
  journal = {arXiv preprint arXiv:2507.18071},
  year = {2025},
  eprint = {2507.18071},
  archivePrefix = {arXiv},
  primaryClass = {cs.LG},
  url = {https://arxiv.org/abs/2507.18071}
}

@article{cui2025entropy,
  author = {{Ganqu Cui} and {Yuchen Zhang} and {Jiacheng Chen} and {Lifan Yuan} and {Zhi Wang} and {Yuxin Zuo} and {Haozhan Li} and {Yuchen Fan} and {Huayu Chen} and {Weize Chen} and {Zhiyuan Liu} and {Hao Peng} and {Lei Bai} and {Wanli Ouyang} and {Yu Cheng} and {Bowen Zhou} and {Ning Ding}},
  title = {{The Entropy Mechanism of Reinforcement Learning for Reasoning Language Models}},
  journal = {arXiv preprint arXiv:2505.22617},
  year = {2025},
  eprint = {2505.22617},
  archivePrefix = {arXiv},
  primaryClass = {cs.LG},
  url = {https://arxiv.org/abs/2505.22617}
}

@article{mroueh2025grpo,
  author = {{Youssef Mroueh}},
  title = {{Reinforcement Learning with Verifiable Rewards: GRPO's Effective Loss, Dynamics, and Success Amplification}},
  journal = {arXiv preprint arXiv:2503.06639},
  year = {2025},
  eprint = {2503.06639},
  archivePrefix = {arXiv},
  primaryClass = {cs.LG},
  url = {https://arxiv.org/abs/2503.06639}
}

@article{yue2025does,
  author = {{Yang Yue} and {Zhiqi Chen} and {Rui Lu} and {Andrew Zhao} and {Zhaokai Wang} and {Yang Yue} and {Shiji Song} and {Gao Huang}},
  title = {{Does Reinforcement Learning Really Incentivize Reasoning Capacity in LLMs Beyond the Base Model?}},
  journal = {arXiv preprint arXiv:2504.13837},
  year = {2025},
  eprint = {2504.13837},
  archivePrefix = {arXiv},
  primaryClass = {cs.AI},
  url = {https://arxiv.org/abs/2504.13837}
}

@article{shao2025spurious,
  author = {{Rulin Shao} and {Shuyue Stella Li} and {Rui Xin} and {Scott Geng} and {Yiping Wang} and {Sewoong Oh} and {Simon Shaolei Du} and {Nathan Lambert} and {Sewon Min} and {Ranjay Krishna} and {Yulia Tsvetkov} and {Hannaneh Hajishirzi} and {Pang Wei Koh} and {Luke Zettlemoyer}},
  title = {{Spurious Rewards: Rethinking Training Signals in RLVR}},
  journal = {arXiv preprint arXiv:2506.10947},
  year = {2025},
  eprint = {2506.10947},
  archivePrefix = {arXiv},
  primaryClass = {cs.AI},
  url = {https://arxiv.org/abs/2506.10947}
}

@article{liu2025prorl,
  author = {{Mingjie Liu} and {Shizhe Diao} and {Ximing Lu} and {Jian Hu} and {Xin Dong} and {Yejin Choi} and {Jan Kautz} and {Yi Dong}},
  title = {{ProRL: Prolonged Reinforcement Learning Expands Reasoning Boundaries in Large Language Models}},
  journal = {arXiv preprint arXiv:2505.24864},
  year = {2025},
  eprint = {2505.24864},
  archivePrefix = {arXiv},
  primaryClass = {cs.CL},
  url = {https://arxiv.org/abs/2505.24864}
}

@article{zeng2025cures,
  author = {{Yongcheng Zeng} and {Zexu Sun} and {Bokai Ji} and {Erxue Min} and {Hengyi Cai} and {Shuaiqiang Wang} and {Dawei Yin} and {Haifeng Zhang} and {Xu Chen} and {Jun Wang}},
  title = {{CurES: From Gradient Analysis to Efficient Curriculum Learning for Reasoning LLMs}},
  journal = {arXiv preprint arXiv:2510.01037},
  year = {2025},
  eprint = {2510.01037},
  archivePrefix = {arXiv},
  primaryClass = {cs.LG},
  url = {https://arxiv.org/abs/2510.01037}
}

@article{chen2025sec,
  author = {{Xiaoyin Chen} and {Jiarui Lu} and {Minsu Kim} and {Dinghuai Zhang} and {Jian Tang} and {Alexandre Pich{\'e}} and {Nicolas Gontier} and {Yoshua Bengio} and {Ehsan Kamalloo}},
  title = {{Self-Evolving Curriculum for LLM Reasoning}},
  journal = {arXiv preprint arXiv:2505.14970},
  year = {2025},
  eprint = {2505.14970},
  archivePrefix = {arXiv},
  primaryClass = {cs.AI},
  url = {https://arxiv.org/abs/2505.14970}
}

@article{foster2025learnability,
  author = {{Thomas Foster} and {Anya Sims} and {Johannes Forkel} and {Mattie Fellows} and {Jakob Foerster}},
  title = {{Learning to Reason at the Frontier of Learnability}},
  journal = {arXiv preprint arXiv:2502.12272},
  year = {2025},
  eprint = {2502.12272},
  archivePrefix = {arXiv},
  primaryClass = {cs.LG},
  url = {https://arxiv.org/abs/2502.12272}
}

@article{parashar2025curriculum,
  author = {{Shubham Parashar} and {Shurui Gui} and {Xiner Li} and {Hongyi Ling} and {Sushil Vemuri} and {Blake Olson} and {Eric Li} and {Yu Zhang} and {James Caverlee} and {Dileep Kalathil} and {Shuiwang Ji}},
  title = {{Curriculum Reinforcement Learning from Easy to Hard Tasks Improves LLM Reasoning}},
  journal = {arXiv preprint arXiv:2506.06632},
  year = {2025},
  eprint = {2506.06632},
  archivePrefix = {arXiv},
  primaryClass = {cs.LG},
  url = {https://arxiv.org/abs/2506.06632}
}

@article{bae2025online,
  author = {{Sanghwan Bae} and {Jiwoo Hong} and {Min Young Lee} and {Hanbyul Kim} and {JeongYeon Nam} and {Donghyun Kwak}},
  title = {{Online Difficulty Filtering for Reasoning Oriented Reinforcement Learning}},
  journal = {arXiv preprint arXiv:2504.03380},
  year = {2025},
  eprint = {2504.03380},
  archivePrefix = {arXiv},
  primaryClass = {cs.CL},
  url = {https://arxiv.org/abs/2504.03380}
}

@inproceedings{mao2026dps,
  author = {{Yixiu Mao} and {Yun Qu} and {Qi Wang} and {Heming Zou} and {Xiangyang Ji}},
  title = {{Dynamics-Predictive Sampling for Active {RL} Finetuning of Large Reasoning Models}},
  booktitle = {International Conference on Learning Representations},
  year = {2026},
  volume = {2026},
  pages = {25242--25270}
}

@article{qu2026gps,
  author = {{Yun Qu} and {Qi Wang} and {Yixiu Mao} and {Heming Zou} and {Yuhang Jiang} and {Weijie Liu} and {Clive Bai} and {Kai Yang} and {Yangkun Chen} and {Saiyong Yang} and {Xiangyang Ji}},
  title = {{Small Generalizable Prompt Predictive Models Can Steer Efficient RL Post-Training of Large Reasoning Models}},
  journal = {arXiv preprint arXiv:2602.01970},
  year = {2026},
  eprint = {2602.01970},
  archivePrefix = {arXiv},
  primaryClass = {cs.AI},
  url = {https://arxiv.org/abs/2602.01970}
}

@article{zhu2026kgps,
  author = {{Haodong Zhu} and {Yangyang Ren} and {Yanjing Li} and {Sheng Xu} and {Haiguang Liu} and {Linlin Yang} and {Baochang Zhang}},
  title = {{Kalman Meets Curriculum: Efficient Dynamic Prompt Selection for Adaptive RL Finetuning}},
  journal = {arXiv preprint arXiv:2607.27610},
  year = {2026},
  eprint = {2607.27610},
  archivePrefix = {arXiv},
  primaryClass = {cs.LG},
  url = {https://arxiv.org/abs/2607.27610}
}

@article{liu2025explore,
  author = {{Chenxi Liu} and {Junjie Liang} and {Yuqi Jia} and {Bochuan Cao} and {Yang Bai} and {Heng Huang} and {Xun Chen}},
  title = {{Explore Data Left Behind in Reinforcement Learning for Reasoning Language Models}},
  journal = {arXiv preprint arXiv:2511.04800},
  year = {2025},
  eprint = {2511.04800},
  archivePrefix = {arXiv},
  primaryClass = {cs.CL},
  url = {https://arxiv.org/abs/2511.04800}
}

@article{mao2026popo,
  author = {{Yixiu Mao} and {Yun Qu} and {Qi Wang} and {Heming Zou} and {Xiangyang Ji}},
  title = {{RLVR without Ineffective Samples: Group Prioritized Off-Policy Optimization for LLM Reasoning}},
  journal = {arXiv preprint arXiv:2606.01281},
  year = {2026},
  eprint = {2606.01281},
  archivePrefix = {arXiv},
  primaryClass = {cs.LG},
  url = {https://arxiv.org/abs/2606.01281}
}

@article{zou2026trace,
  author = {{Heming Zou} and {Qi Wang} and {Yun Qu} and {Yuhang Jiang} and {Lizhou Cai} and {Yixiu Mao} and {Ru Peng} and {Xin Xu} and {Weijie Liu} and {Kai Yang} and {Saiyong Yang} and {Xiangyang Ji}},
  title = {{{TRACE}: A Unified Rollout Budget Allocation Framework for Efficient Agentic Reinforcement Learning}},
  journal = {arXiv preprint arXiv:2606.11119},
  year = {2026},
  eprint = {2606.11119},
  archivePrefix = {arXiv},
  primaryClass = {cs.LG},
  url = {https://arxiv.org/abs/2606.11119}
}

@article{baroian2026replay,
  author = {{Andrei Baroian} and {Rutger Berger}},
  title = {{Prompt replay: speeding up grpo with on-policy reuse of high-signal prompts}},
  journal = {arXiv preprint arXiv:2603.21177},
  year = {2026},
  eprint = {2603.21177},
  archivePrefix = {arXiv},
  primaryClass = {cs.LG},
  url = {https://arxiv.org/abs/2603.21177}
}

@article{li2025limr,
  author = {{Xuefeng Li} and {Haoyang Zou} and {Pengfei Liu}},
  title = {{LIMR: Less is More for RL Scaling}},
  journal = {arXiv preprint arXiv:2502.11886},
  year = {2025},
  eprint = {2502.11886},
  archivePrefix = {arXiv},
  primaryClass = {cs.LG},
  url = {https://arxiv.org/abs/2502.11886}
}

@article{zeng2025eachprompt,
  author = {{Anxiang Zeng} and {Haibo Zhang} and {Hailing Zhang} and {Kaixiang Mo} and {Liang Yao} and {Ling Hu} and {Long Zhang} and {Shuman Liu} and {Shuyi Xie} and {Yanshi Li} and {Yizhang Chen} and {Yuepeng Sheng} and {Yuwei Huang} and {Zhaochen Xu} and {Zhiqiang Zhou} and {Ziqin Liew}},
  title = {{Each Prompt Matters: Scaling Reinforcement Learning Without Wasting Rollouts on Hundred-Billion-Scale MoE}},
  journal = {arXiv preprint arXiv:2512.07710},
  year = {2025},
  eprint = {2512.07710},
  archivePrefix = {arXiv},
  primaryClass = {cs.AI},
  url = {https://arxiv.org/abs/2512.07710}
}

@article{liang2026leeps,
  author = {{Shuang Liang} and {Haoyang Zhou} and {Yifan Gong} and {Guowei Wang} and {Xiting Wang}},
  title = {{LEEPS: Latent-Guided Explore-Exploit Prompt Sampling for Efficient RLVR in Large Language Models}},
  journal = {arXiv preprint arXiv:2607.28077},
  year = {2026},
  eprint = {2607.28077},
  archivePrefix = {arXiv},
  primaryClass = {cs.CL},
  url = {https://arxiv.org/abs/2607.28077}
}

@article{sha2026thinkprior,
  author = {{Tommy Sha} and {Skylar Zhai} and {Siqi Zhao}},
  title = {{ThinkPrior: Zero-Rollout Difficulty Priors for Cold-Start Prompt Selection in RLVR}},
  journal = {arXiv preprint arXiv:2609.09075},
  year = {2026},
  eprint = {2609.09075},
  archivePrefix = {arXiv},
  primaryClass = {cs.LG},
  url = {https://arxiv.org/abs/2609.09075}
}

@article{shi2025adarft,
  author = {{Taiwei Shi} and {Yiyang Wu} and {Linxin Song} and {Tianyi Zhou} and {Jieyu Zhao}},
  title = {{Efficient Reinforcement Finetuning via Adaptive Curriculum Learning}},
  journal = {arXiv preprint arXiv:2504.05520},
  year = {2025},
  eprint = {2504.05520},
  archivePrefix = {arXiv},
  primaryClass = {cs.LG},
  url = {https://arxiv.org/abs/2504.05520}
}

@article{schulman2017ppo,
  author = {{John Schulman} and {Filip Wolski} and {Prafulla Dhariwal} and {Alec Radford} and {Oleg Klimov}},
  title = {{Proximal Policy Optimization Algorithms}},
  journal = {arXiv preprint arXiv:1707.06347},
  year = {2017},
  eprint = {1707.06347},
  archivePrefix = {arXiv},
  primaryClass = {cs.LG},
  url = {https://arxiv.org/abs/1707.06347}
}

@article{yang2025qwen3,
  author = {{An Yang} and {Anfeng Li} and {Baosong Yang} and {Beichen Zhang} and {Binyuan Hui} and {Bo Zheng} and {Bowen Yu} and {Chang Gao} and {Chengen Huang} and {Chenxu Lv} and {Chujie Zheng} and {Dayiheng Liu} and {Fan Zhou} and {Fei Huang} and {Feng Hu} and {Hao Ge} and {Haoran Wei} and {Huan Lin} and {Jialong Tang} and {Jian Yang} and {Jianhong Tu} and {Jianwei Zhang} and {Jianxin Yang} and {Jiaxi Yang} and {Jing Zhou} and {Jingren Zhou} and {Junyang Lin} and {Kai Dang} and {Keqin Bao} and {Kexin Yang} and {Le Yu} and {Lianghao Deng} and {Mei Li} and {Mingfeng Xue} and {Mingze Li} and {Pei Zhang} and {Peng Wang} and {Qin Zhu} and {Rui Men} and {Ruize Gao} and {Shixuan Liu} and {Shuang Luo} and {Tianhao Li} and {Tianyi Tang} and {Wenbiao Yin} and {Xingzhang Ren} and {Xinyu Wang} and {Xinyu Zhang} and {Xuancheng Ren} and {Yang Fan} and {Yang Su} and {Yichang Zhang} and {Yinger Zhang} and {Yu Wan} and {Yuqiong Liu} and {Zekun Wang} and {Zeyu Cui} and {Zhenru Zhang} and {Zhipeng Zhou} and {Zihan Qiu}},
  title = {{Qwen3 Technical Report}},
  journal = {arXiv preprint arXiv:2505.09388},
  year = {2025},
  eprint = {2505.09388},
  archivePrefix = {arXiv},
  primaryClass = {cs.CL},
  url = {https://arxiv.org/abs/2505.09388}
}

@article{qwen2024qwen25,
  author = {{Qwen} and {An Yang} and {Baosong Yang} and {Beichen Zhang} and {Binyuan Hui} and {Bo Zheng} and {Bowen Yu} and {Chengyuan Li} and {Dayiheng Liu} and {Fei Huang} and {Haoran Wei} and {Huan Lin} and {Jian Yang} and {Jianhong Tu} and {Jianwei Zhang} and {Jianxin Yang} and {Jiaxi Yang} and {Jingren Zhou} and {Junyang Lin} and {Kai Dang} and {Keming Lu} and {Keqin Bao} and {Kexin Yang} and {Le Yu} and {Mei Li} and {Mingfeng Xue} and {Pei Zhang} and {Qin Zhu} and {Rui Men} and {Runji Lin} and {Tianhao Li} and {Tianyi Tang} and {Tingyu Xia} and {Xingzhang Ren} and {Xuancheng Ren} and {Yang Fan} and {Yang Su} and {Yichang Zhang} and {Yu Wan} and {Yuqiong Liu} and {Zeyu Cui} and {Zhenru Zhang} and {Zihan Qiu}},
  title = {{Qwen2.5 Technical Report}},
  journal = {arXiv preprint arXiv:2412.15115},
  year = {2024},
  eprint = {2412.15115},
  archivePrefix = {arXiv},
  primaryClass = {cs.CL},
  url = {https://arxiv.org/abs/2412.15115}
}

@article{cobbe2021gsm8k,
  author = {{Karl Cobbe} and {Vineet Kosaraju} and {Mohammad Bavarian} and {Mark Chen} and {Heewoo Jun} and {Lukasz Kaiser} and {Matthias Plappert} and {Jerry Tworek} and {Jacob Hilton} and {Reiichiro Nakano} and {Christopher Hesse} and {John Schulman}},
  title = {{Training Verifiers to Solve Math Word Problems}},
  journal = {arXiv preprint arXiv:2110.14168},
  year = {2021},
  eprint = {2110.14168},
  archivePrefix = {arXiv},
  primaryClass = {cs.LG},
  url = {https://arxiv.org/abs/2110.14168}
}

@article{hendrycks2021math,
  author = {{Dan Hendrycks} and {Collin Burns} and {Saurav Kadavath} and {Akul Arora} and {Steven Basart} and {Eric Tang} and {Dawn Song} and {Jacob Steinhardt}},
  title = {{Measuring Mathematical Problem Solving With the MATH Dataset}},
  journal = {arXiv preprint arXiv:2103.03874},
  year = {2021},
  eprint = {2103.03874},
  archivePrefix = {arXiv},
  primaryClass = {cs.LG},
  url = {https://arxiv.org/abs/2103.03874}
}

@article{lightman2023verify,
  author = {{Hunter Lightman} and {Vineet Kosaraju} and {Yura Burda} and {Harri Edwards} and {Bowen Baker} and {Teddy Lee} and {Jan Leike} and {John Schulman} and {Ilya Sutskever} and {Karl Cobbe}},
  title = {{Let's Verify Step by Step}},
  journal = {arXiv preprint arXiv:2305.20050},
  year = {2023},
  eprint = {2305.20050},
  archivePrefix = {arXiv},
  primaryClass = {cs.LG},
  url = {https://arxiv.org/abs/2305.20050}
}

@article{lewkowycz2022minerva,
  author = {{Aitor Lewkowycz} and {Anders Andreassen} and {David Dohan} and {Ethan Dyer} and {Henryk Michalewski} and {Vinay Ramasesh} and {Ambrose Slone} and {Cem Anil} and {Imanol Schlag} and {Theo Gutman-Solo} and {Yuhuai Wu} and {Behnam Neyshabur} and {Guy Gur-Ari} and {Vedant Misra}},
  title = {{Solving Quantitative Reasoning Problems with Language Models}},
  journal = {arXiv preprint arXiv:2206.14858},
  year = {2022},
  eprint = {2206.14858},
  archivePrefix = {arXiv},
  primaryClass = {cs.CL},
  url = {https://arxiv.org/abs/2206.14858}
}

@article{he2024olympiadbench,
  author = {{Chaoqun He} and {Renjie Luo} and {Yuzhuo Bai} and {Shengding Hu} and {Zhen Leng Thai} and {Junhao Shen} and {Jinyi Hu} and {Xu Han} and {Yujie Huang} and {Yuxiang Zhang} and {Jie Liu} and {Lei Qi} and {Zhiyuan Liu} and {Maosong Sun}},
  title = {{OlympiadBench: A Challenging Benchmark for Promoting AGI with Olympiad-Level Bilingual Multimodal Scientific Problems}},
  journal = {arXiv preprint arXiv:2402.14008},
  year = {2024},
  eprint = {2402.14008},
  archivePrefix = {arXiv},
  primaryClass = {cs.CL},
  url = {https://arxiv.org/abs/2402.14008}
}

@article{yang2018hotpotqa,
  author = {{Zhilin Yang} and {Peng Qi} and {Saizheng Zhang} and {Yoshua Bengio} and {William W. Cohen} and {Ruslan Salakhutdinov} and {Christopher D. Manning}},
  title = {{HotpotQA: A Dataset for Diverse, Explainable Multi-hop Question Answering}},
  journal = {arXiv preprint arXiv:1809.09600},
  year = {2018},
  eprint = {1809.09600},
  archivePrefix = {arXiv},
  primaryClass = {cs.CL},
  url = {https://arxiv.org/abs/1809.09600}
}

@article{trivedi2022musique,
  author = {{Harsh Trivedi} and {Niranjan Balasubramanian} and {Tushar Khot} and {Ashish Sabharwal}},
  title = {{MuSiQue: Multihop Questions via Single-hop Question Composition}},
  journal = {arXiv preprint arXiv:2108.00573},
  year = {2021},
  eprint = {2108.00573},
  archivePrefix = {arXiv},
  primaryClass = {cs.CL},
  url = {https://arxiv.org/abs/2108.00573}
}

@article{ho2020twowiki,
  author = {{Xanh Ho} and {Anh-Khoa Duong Nguyen} and {Saku Sugawara} and {Akiko Aizawa}},
  title = {{Constructing A Multi-hop QA Dataset for Comprehensive Evaluation of Reasoning Steps}},
  journal = {arXiv preprint arXiv:2011.01060},
  year = {2020},
  eprint = {2011.01060},
  archivePrefix = {arXiv},
  primaryClass = {cs.CL},
  url = {https://arxiv.org/abs/2011.01060}
}

@article{press2023bamboogle,
  author = {{Ofir Press} and {Muru Zhang} and {Sewon Min} and {Ludwig Schmidt} and {Noah A. Smith} and {Mike Lewis}},
  title = {{Measuring and Narrowing the Compositionality Gap in Language Models}},
  journal = {arXiv preprint arXiv:2210.03350},
  year = {2022},
  eprint = {2210.03350},
  archivePrefix = {arXiv},
  primaryClass = {cs.CL},
  url = {https://arxiv.org/abs/2210.03350}
}

@article{karpukhin2020dpr,
  author = {{Vladimir Karpukhin} and {Barlas O{\u{g}}uz} and {Sewon Min} and {Patrick Lewis} and {Ledell Wu} and {Sergey Edunov} and {Danqi Chen} and {Wen-tau Yih}},
  title = {{Dense Passage Retrieval for Open-Domain Question Answering}},
  journal = {arXiv preprint arXiv:2004.04906},
  year = {2020},
  eprint = {2004.04906},
  archivePrefix = {arXiv},
  primaryClass = {cs.CL},
  url = {https://arxiv.org/abs/2004.04906}
}

@article{sheng2024hybridflow,
  author = {{Guangming Sheng} and {Chi Zhang} and {Zilingfeng Ye} and {Xibin Wu} and {Wang Zhang} and {Ru Zhang} and {Yanghua Peng} and {Haibin Lin} and {Chuan Wu}},
  title = {{HybridFlow: A Flexible and Efficient RLHF Framework}},
  journal = {arXiv preprint arXiv:2409.19256},
  year = {2024},
  eprint = {2409.19256},
  archivePrefix = {arXiv},
  primaryClass = {cs.LG},
  url = {https://arxiv.org/abs/2409.19256}
}

@article{kwon2023vllm,
  author = {{Woosuk Kwon} and {Zhuohan Li} and {Siyuan Zhuang} and {Ying Sheng} and {Lianmin Zheng} and {Cody Hao Yu} and {Joseph E. Gonzalez} and {Hao Zhang} and {Ion Stoica}},
  title = {{Efficient Memory Management for Large Language Model Serving with PagedAttention}},
  journal = {arXiv preprint arXiv:2309.06180},
  year = {2023},
  eprint = {2309.06180},
  archivePrefix = {arXiv},
  primaryClass = {cs.LG},
  url = {https://arxiv.org/abs/2309.06180}
}

\end{document}